\documentclass{article}

\PassOptionsToPackage{numbers,sort&compress}{natbib}
\usepackage[preprint]{neurips_2026}

\usepackage[T1]{fontenc}
\usepackage[utf8]{inputenc}
\usepackage{microtype}
\usepackage{amsmath,amssymb,amsthm}
\usepackage{graphicx}
\usepackage{xcolor}
\usepackage{url}
\usepackage{comment}
\usepackage[colorlinks=true, linkcolor=blue!60!black, citecolor=blue!60!black,
            urlcolor=blue!60!black]{hyperref}
\hypersetup{
  pdftitle={The Dynamics of Discoverability: How Trajectories and Priors Shape Equation Recovery},
  pdfauthor={Matteo Gallo, Fabio Anselmi, Paolo Lazzari}
}

\numberwithin{equation}{section}

\title{The Dynamics of Discoverability:\\How Trajectories and Priors\\Shape Equation Recovery}

\author{%
  Matteo Gallo\textsuperscript{1,2}\thanks{Corresponding author: \texttt{matteo.gallo@phd.units.it}} \qquad
  Fabio Anselmi\textsuperscript{1} \qquad
  Paolo Lazzari\textsuperscript{2,3} \\[3pt]
  \textsuperscript{1}University of Trieste, Trieste, Italy \\
  \textsuperscript{2}National Institute of Oceanography and Applied Geophysics - OGS, Trieste, Italy \\
  \textsuperscript{3}NBFC, National Biodiversity Future Center, Palermo, Italy \\
}

\begin{document}
\maketitle

\begin{abstract}
How the dynamical regime of the observed system affects equation discovery has
mainly been investigated through comparisons across systems. However, such comparisons vary
both the equations and the dynamics, confounding the effect of the regime with the
difficulty of recovering the equations symbolically. We separate the two by varying
the forcing of Lorenz-84, moving its fully observed post-transient
trajectories through fixed-point, periodic, and chaotic regimes while preserving the
equations' functional form. Within each regime, we separately vary the amount of data,
the noise, and prior knowledge of which terms the equations contain. We then measure
how well two complementary approaches, sparse regression over a fixed library of
candidate functions (SINDy) and an evolutionary search over symbolic expression trees (PySR), recover
the true equations' terms and coefficients. We find that recovery depends on whether
the sampled states distinguish combinations of candidate functions: equations remain
poorly recovered from fixed-point data even when the candidate set contains only the
true terms. We link the effect of the dynamics on both algorithms to one object: the
moment matrix of the candidate functions under the invariant measure of the regime.
Small eigenvalues mark weakly distinguishable combinations of candidate functions: we
show that more data, less noise, and more prior knowledge can mitigate the resulting
recovery difficulties, while a zero eigenvalue makes distinct equations indistinguishable
on the visited states. Hence, a more precise prior needs less informative data,
with consequences for data collection, method design, and evaluation.
\end{abstract}

%

\section{Introduction}\label{sec:introduction}

Discovering governing equations from observed trajectories is a central problem in
data-driven science. Prominent methods include sparse identification of nonlinear dynamics
(SINDy)~\citep{brunton2016discovering} and evolutionary symbolic regression with
PySR~\citep{cranmer2023pysr,schmidt2009distilling}, alongside neural
approaches~\citep{srbench2024}. These methods
are routinely benchmarked on chaotic systems, a demanding test
case~\citep{brunton2016discovering,kaptanoglu2023benchmarking,gilpin2021chaos}, while
recent theory argues that discovery requires chaos~\citep{discoverability2025}. Is
chaos an obstacle to equation discovery, or a requirement? The dynamical regime
matters because it determines which states are observed and how often: if two
candidate equations behave similarly on the visited states, the data cannot tell
them apart. We therefore ask how recovery changes across regimes.

Earlier work asks when observations can identify a governing equation. System
identification studies how well data constrain an assumed model, using
persistent excitation and Fisher information~\citep{markovsky2022data,
astrom2008adaptive,bao2025fisher}. If the unknown vector field can be any
continuous function, one trajectory can identify it across a target region only if
it visits every part of that region~\citep{discoverability2025}.
Other work assumes that the equation uses only a few functions from a specified
set. With that restriction, recovery can be proved under particular sampling
assumptions, including cases with relatively few observations, noisy
measurements, or outliers~\citep{tran2017exact,schaeffer2018extreme,
schaeffer2020extracting,ho2018recovery}. However, the quantitative effect of dynamical
regime is difficult to isolate empirically. Existing benchmarks compare different systems 
with different amounts of chaos, rather than changing the regime of 
one fixed system~\citep{kaptanoglu2023benchmarking,
gilpin2021chaos}. Their governing equations therefore also change: they contain
different terms, differ in complexity, and require different operators.
These studies show no consistent relationship between the amount
of chaos and recovery performance, but because the equations change with the
dynamics, they do not isolate the effect of the dynamics on recovery.
\citet{discoverability2025} vary the forcing of a single system, but assess a
ridge-regression fit by its out-of-domain prediction error and posterior
uncertainty, not by symbolic recovery, which prediction error does not
establish~\citep{matsubara2022rethinking}.

We isolate the dynamical contribution by designing experiments where we choose a fixed target system,
the Lorenz-84 atmospheric model~\citep{lorenz1984irregularity}, 
and vary only one forcing parameter to
produce fixed-point, limit-cycle, and chaotic trajectories without changing the governing 
functional form.
We choose SINDy and PySR because they share no internal mechanism, so a dependence
on the regime common to both can be attributed to the data rather than to either
algorithm. To separate the dynamics from the other variables of equation discovery,
we vary training-set size, measurement noise, and prior information separately
within each regime.
To measure partial recovery, we introduce Soft~F1, which scores
both the recovered terms and the accuracy of their coefficients.
We then evaluate the Lorenz-84 results without modification on the held-out Lorenz-96
system~\citep{lorenz1996predictability} and on three polynomial and
non-polynomial jerk systems~\citep{sprott1997some}.

We make three contributions. First, we find that, for a fixed candidate library,
chaos is neither an obstacle nor a requirement: what matters is whether the
trajectory visits states that separate the candidate equations the prior leaves
open, and the fewer it leaves open, the less the trajectory must explore. With the
oracle prior, even a weakly excited limit cycle is recovered perfectly, whereas
fixed-point data remain poorly recovered even then. We also find that equations
containing none of the true terms can predict held-out data perfectly, so
predictive accuracy alone cannot establish discovery. Second, we connect a known
object, the moment matrix of the candidate functions under the regime's invariant
measure, to how the two algorithms fail. This matrix is used in system 
identification~\citep{markovsky2022data} and by
\citet{discoverability2025} to decide whether data can identify a model; 
here its eigenvalues measure how distinguishable each combination of
candidate functions is, and indicators built from each algorithm's failure
mechanism, including PySR's evolutionary search, support this link on Lorenz-84 and
on the held-out systems. Its smallest eigenvalue can be computed from the data
before running discovery, as a cheap check of whether they separate the candidate
functions. Third, with experiments and an ideal-observer analysis that places the 
regime, data amount, noise, and prior on a common scale, we show how they combine:
data, noise, and prior can mitigate weak distinguishability, but no amount of data
separates equations that produce identical data on the visited states.

\section{Background and Related Work}\label{sec:background}

\subsection{Sparse identification of nonlinear dynamics}
SINDy~\citep{brunton2016discovering} represents each state derivative as a sparse
linear combination of candidate functions. Given $N$ state observations, it evaluates
$p$ candidate functions to form $\Theta\in\mathbb{R}^{N\times p}$ and solves
$\dot X\approx\Theta\xi$, where the nonzero entries of
$\xi\in\mathbb{R}^{p\times d}$ identify the active terms and their coefficients.
Sequentially thresholded least squares alternates ridge regression with hard
thresholding until the active set stabilizes. Its principal hyperparameters are the
ridge penalty and sparsity threshold.

\subsection{Evolutionary symbolic regression}
PySR~\citep{cranmer2023pysr} searches expression trees assembled from primitive
operators rather than restricting the model to linear combinations of a fixed
library.
PySR evolves a population of expressions by residual loss and
complexity, locally optimizes their numerical coefficients using
BFGS~\citep{broyden1970convergence}, and retains a Pareto front from which the final
expression is selected. The fixed operators, search budget, complexity limits, and
selection rule used in our experiments are reported in
Appendix~\ref{app:experimental-details}.

To avoid confusion in the terminology used below: SINDy has a finite 
library of candidate functions. PySR has an operator
set. Recursive combinations of these operators define a combinatorial expression space.
Hence, in this paper, \emph{library} is always related to SINDy, while \emph{operator set} to PySR.
We use \emph{dictionary} for the finite collection of candidate functions $\Phi$ 
evaluated in the common moment matrix analysis.

\subsection{Lorenz-84 controlled benchmark system}
The Lorenz-84 system~\citep{lorenz1984irregularity},
\begin{align*}
  \dot{x}_0 &= -x_1^2 - x_2^2 - Ax_0 + AF,\\
  \dot{x}_1 &= \phantom{-}x_0 x_1 - Bx_0 x_2 - x_1 + G,\\
  \dot{x}_2 &= \phantom{-}Bx_0 x_1 + x_0 x_2 - x_2,
\end{align*}
is a three-dimensional polynomial model of large-scale atmospheric dynamics. We fix
$A=0.25$, $B=4$, and $G=1$ and vary only the forcing $F$, producing fixed-point,
limit-cycle, and chaotic trajectories without changing the governing functional
form. This makes L84 the controlled system for isolating the effect of dynamical
regime on recovery.

\subsection{Lorenz-96 held-out validation system}
The five-dimensional Lorenz-96 system~\citep{lorenz1996predictability} is
\[
  \dot{x}_i=(x_{i+1}-x_{i-2})x_{i-1}-x_i+F,
  \qquad i=1,\ldots,5,
\]
with cyclic indices. Varying $F$ again produces fixed-point, limit-cycle, and chaotic
regimes, but with a different dimension and interaction structure. We use L96 as a
held-out test of whether the quantities developed on L84 transfer without refitting.

\subsection{Soft~F1 performance metric}
Predictive metrics such as RMSE and $R^2$ assess the agreement on observed data but do
not establish that the recovered symbolic terms are correct. The Binary structural F1
distinguishes present and absent terms, but gives equal credit to accurately and
inaccurately estimated coefficients. We introduce Soft~F1, a metric that measures
symbolic term and coefficient recovery. This is useful
when noise, limited data, or an imperfect candidate class produce intermediate
rather than exactly correct expressions~\citep{delahunt2022toolkit,cortiella2020sparse}.

Let $\mathcal T$ and $\mathcal D$ be the ground-truth and discovered term sets.
For each $j\in\mathcal T\cup\mathcal D$, with a missing
coefficient set to zero, define
\begin{equation}
  s_j=\exp(-\alpha\delta_j),
  \qquad
  \delta_j=\frac{|c_j-\hat c_j|}
  {|c_j|+|\hat c_j|+\epsilon},
  \label{eq:softf1-fidelity}
\end{equation}
where $\epsilon=10^{-12}$ and we use $\alpha=3$. Soft precision and recall average
this fidelity over discovered and ground-truth terms, respectively:
\begin{equation}
  P_{\mathrm{soft}}=\frac{1}{|\mathcal D|}
  \sum_{j\in\mathcal D}s_j,
  \qquad
  R_{\mathrm{soft}}=\frac{1}{|\mathcal T|}
  \sum_{j\in\mathcal T}s_j,
  \qquad
  \mathrm{Soft~F1}
  =\frac{2P_{\mathrm{soft}}R_{\mathrm{soft}}}
  {P_{\mathrm{soft}}+R_{\mathrm{soft}}}.
  \label{eq:softf1}
\end{equation}
We compute Soft~F1
separately for each state equation and average the resulting scores across state
dimensions. It complements structural F1, predictive metrics,
and expression-tree distances such as NED and TED~\citep{matsubara2022rethinking,
reis2024benchmarking}. Appendix~\ref{si:softf1-definition} gives more details.

\section{Results}\label{sec:results}

\subsection{Dynamical regime is consistently associated with equation-recovery difficulty}

We apply SINDy and PySR to eight regimes produced by Lorenz-84: one fixed point (R0), two limit-cycle
branches differentiated by attractor amplitude (R1--R2), and five chaotic regimes ordered
by the maximum Lyapunov exponent (R3--R7). Limit-cycle candidates are additionally
verified using the second Lyapunov exponent. We then vary three factors independently:
17 noiseless training-set sizes log-spaced from $N=20$ to $10{,}000$; seven state-noise ratios ($\eta$)
from $0$ to $20\%$, with finite-difference (FD) or Savitzky--Golay (SG) derivative estimation;
and three levels of structural prior information at fixed $N=5{,}000$:
null (ground-truth terms excluded), overcomplete (ground-truth terms
available alongside wrong ones), and oracle (only ground-truth terms
available). SINDy uses five independent trajectory slots per regime, while PySR uses
five slots and five algorithm initializations. Recovery is measured by
Soft~F1 (Eq.~\ref{eq:softf1}).

\begin{figure}[t!]
\centering
\includegraphics[width=\linewidth]{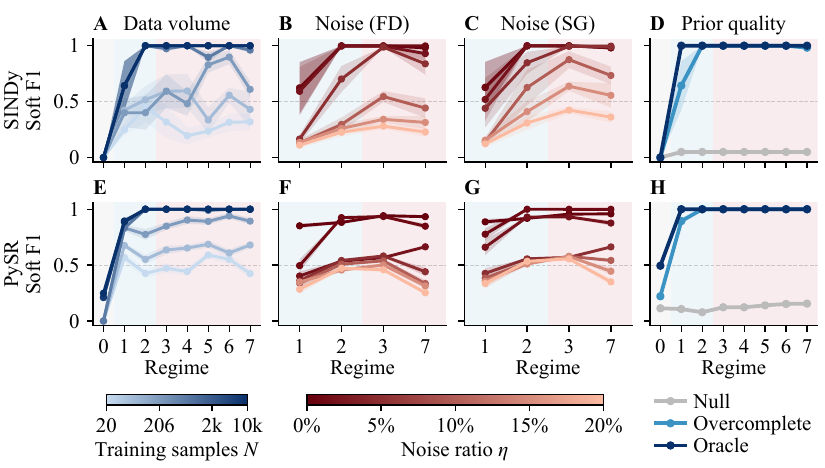}
\caption{\textbf{Dynamical regime is consistently associated with
equation-recovery difficulty, although the precise effect is not universal.}
Horizontal axes give the index $k$ of regime R$k$; grey, blue, and red backgrounds identify the fixed
point, limit cycles, and chaotic regimes. Panels A and E vary the noiseless
training size; B, C, F, and G vary measurement noise using finite differences (FD)
or Savitzky--Golay smoothing (SG); D and H vary prior information quantity. Soft~F1 is
computed per state equation and averaged within each fit. Curves show means
over independent fits (SINDy: five trajectory slots; PySR: five slots and up to
five initializations), with shaded bands denoting $\pm1$ s.e.m.}
\label{fig:f2}
\end{figure}

In data volume experiments (Fig.~\ref{fig:f2}A,E), R0 is poorly recovered by both 
algorithms at every $N$ (Soft~F1$\leq0.25$). The low-amplitude limit cycle R1 slowly improves
and plateaus at $\approx0.64$ for SINDy and $\approx0.89$ for PySR, whereas R2 
and the chaotic regimes reach Soft~F1$\geq0.99$ by $N\approx2{,}100$.
At $N=20$, every regime scores above R0, but all remain far below perfect recovery.
From $N\approx660$ onward, R2 and the chaotic regimes are recovered substantially 
better than R0 and R1. However, among R2--R7, no consistent ordering emerges: 
below $N\approx2{,}100$ the ranking changes with $N$ and algorithm, and the chaotic 
regimes are neither consistently better nor consistently worse than R2. Additional 
data therefore do not close the gap for R0 and R1, and while we see that richer 
dynamics favour recovery, chaos alone does not determine how well a regime is recovered.

Measurement noise degrades every regime and how fast it depends on the algorithm and 
derivative estimator (Fig.~\ref{fig:f2}B,C,F,G). R1 is the lowest regime in almost 
every condition, but the richer regimes do not follow a fixed order. Under SINDy, chaotic
regimes tolerate noise better than R2 (at $10\%$ with finite differences, R3 and R7 
retain Soft~F1 $0.55$ and $0.44$ against $0.30$ for R2), whereas under PySR with finite
differences, R7 falls to the level of R1 ($0.25$ versus $0.28$ at $20\%$) while R2 and 
R3 stay near $0.47$. Savitzky--Golay smoothing moves these failures to higher noise 
without removing them. Richer dynamics therefore improve noise tolerance, but chaos by
itself does not guarantee it, and failures remain specific to the algorithm and preprocessing.

Prior quality changes recovery but cannot compensate for information absent from the 
trajectory (Fig.~\ref{fig:f2}D,H). With overcomplete or oracle priors, R2--R7 are recovered
almost perfectly by both algorithms (Soft~F1$\geq0.98$). R1 depends on the prior, reaching
$0.64$ (SINDy) and $0.89$ (PySR) under the overcomplete prior and $1.0$ under the oracle.
R0 remains poorly recovered even under the oracle prior (Soft~F1 $0$ for SINDy and $0.49$ for
PySR): at a fixed point, the true terms alone can be combined into other equations that fit
the data equally well (Proposition~\ref{prop:spark}). 
Under the null prior, which excludes every 
ground-truth term, Soft~F1 stays at or below $0.16$, yet SINDy reaches a held-out $R^2$
of $1.00$ on two of the five R1 trajectories: a model can predict perfectly with the wrong equation.
Appendix~\ref{app:information-support} shows that such an error can be physical and not
only symbolic. Benchmarks of discovery algorithms should therefore always score recovered 
equations symbolically, not only by predictive accuracy alone.

In the three experiments, R0 and R1 are generally the most difficult regimes to recover, while
R2 and the chaotic regimes do not follow a fixed order. In particular, chaos is neither 
systematically easier nor systematically more difficult than the high-amplitude limit cycle.
Consequently, the dynamical regime influences discoverability but does not determine it. This
suggests that recovery depends on how the dynamical properties of the trajectory interact
with the candidate library, the noise, and the algorithm.

\subsection{Mechanistic indicators explain the regime dependence of discoverability}

We now test this suggestion. For each algorithm, we build a recovery indicator from the
trajectory, the candidate terms, the true equation, and the noise. Each indicator is based
on how its algorithm selects terms. The SINDy indicator follows directly from sparse
regression, with no free choices. The PySR indicator combines two factors and we choose
how to combine them using L84. To validate both indicators, we then apply them without
changes to L96, which we did not use to build them.

For SINDy, we model the first step of its algorithm: a ridge regression followed by a
threshold that removes small coefficients. The trajectory sets how much the regression
mixes true and wrong library terms, and how much noise affects each coefficient estimate.
From the mean and variance of each coefficient, we compute the probability that it
survives the threshold. Summing these probabilities gives the expected numbers of true
positives (true terms selected), false positives (wrong terms selected), and false
negatives (true terms discarded). We combine these counts in the form of a binary F1
score. This form rewards true terms that are selected, penalizes wrong terms selected and
true terms discarded with equal weight, and ignores wrong terms correctly discarded, which
are the majority in a sparse library and would inflate the score:
\begin{equation}
\mathcal I_{\mathrm{SINDy}}
=
\frac{2\mathbb E[\mathrm{TP}]}
{2\mathbb E[\mathrm{TP}]
+\mathbb E[\mathrm{FP}]
+\mathbb E[\mathrm{FN}]}.
\label{eq:fsindy}
\end{equation}
Without noise, this is the result of SINDy's first step in the training data. It is
therefore not an independent prediction of SINDy, but it shows how the trajectory decides
which terms survive. Appendix~\ref{app:sindy} gives the complete derivation, including how
the noise is propagated through the derivative estimator.

PySR builds expressions using its operator set, without a fixed library. We therefore
cannot model its steps as we did for SINDy. Instead, we measure two properties of the
trajectory. The first, $\sigma_{\mathrm{sep}}$, measures how well the trajectory
separates true terms from wrong ones. It is zero when the wrong terms can exactly reproduce
some combination of true terms along the trajectory, up to a constant. The second,
$Q_{\mathrm{noise}}$, measures whether the part of each true term that the wrong terms
cannot reproduce is larger than the noise. If it is smaller, noise hides the difference,
and PySR can select a wrong term instead. $Q_{\mathrm{noise}}$ equals one without noise.
We combine the two factors as follows
\begin{equation}
\mathcal F_{\mathrm{PySR}}
=
\sqrt{\sigma_{\mathrm{sep}}^{1/4}\,Q_{\mathrm{noise}}}.
\label{eq:psmp}
\end{equation}
We chose this combination only using L84; other reasonable choices change the
correlations with Soft~F1 by at most $0.03$. Appendix~\ref{app:pysr} gives the
construction and comparison with other choices.

Table~\ref{tab:noiseless} shows that both indicators follow Soft~F1 in all three
experiments, with similar correlations on L84 and on the held-out L96. For SINDy, this
means that the first step largely decides the final result. The PySR indicator does not
simulate the algorithm, so its correlation is a more independent test. Without noise,
both indicators depend on the trajectory only through moment matrix quantities: the
SINDy first step through the moment matrix of the true and candidate terms and the
number of samples, and the PySR indicator through $\sigma_{\mathrm{sep}}$, since
$Q_{\mathrm{noise}}=1$. The data volume and prior quality experiments therefore test
this link directly. The same indicators also reproduce the regime dependence in the
non-polynomial jerk systems (Appendix~\ref{app:jerk}). We study the moment matrix and
its link to the invariant measure in the next subsection.

\begin{table}[t!]
\centering
\caption{\textbf{Mechanistic indicators follow Soft~F1 across systems and
experiments.} Entries are Spearman correlations $\rho$ between each indicator and
Soft~F1, computed for each state equation across experimental cells and then averaged
over equations; every per-equation correlation is positive. The last column averages
the three experiments. Appendix~\ref{si:stats-hygiene} gives confidence intervals, computed by
resampling whole trajectories because cells that share a trajectory are not
independent.}
\vspace{+0.3cm}
\begin{tabular}{llcccc}
\hline
Score & System & Data volume & Noise & Prior quality & Average \\
\hline
$\mathcal{I}_{\mathrm{SINDy}}$ & L84 & $0.85$ & $0.92$ & $0.92$ & $0.90$ \\
$\mathcal{I}_{\mathrm{SINDy}}$ & L96 & $0.91$ & $0.90$ & $0.91$ & $0.90$ \\
$\mathcal{F}_{\mathrm{PySR}}$  & L84 & $0.78$ & $0.81$ & $0.86$ & $0.82$ \\
$\mathcal{F}_{\mathrm{PySR}}$  & L96 & $0.82$ & $0.88$ & $0.91$ & $0.87$ \\
\hline
\end{tabular}
\label{tab:noiseless}
\end{table}

\subsection{The moment matrix links the dynamical regime to discoverability}

The two indicators describe how each algorithm uses the data. The data themselves,
however, limit what any algorithm can recover: two candidate equations can be told apart only
if they behave differently on the states that the trajectory visits. Two vector
fields $f$ and $g$ may differ in regions that the trajectory never visits, and the
data cannot show this difference. We therefore compare them only on the visited
states, using the average squared difference
\begin{equation}
D_\mu(f,g)=\int\|f(x)-g(x)\|_2^2\,d\mu(x).
\label{eq:vectorfielddiscrepancy}
\end{equation}
Here $\mu$ is the invariant measure of the regime: it describes how often the
trajectory visits each region of state space. For a finite sample trajectory
$x_1,\dots,x_N$, the integral becomes the average over the samples,
$\frac{1}{N}\sum_{n}\|f(x_n)-g(x_n)\|_2^2$.

If $D_\mu=0$, the two equations give the same derivative in each visited state. They
then generate the same trajectories there, so no method can tell them apart from these
data (Appendix~\ref{app:theory}, Proposition~\ref{prop:equivalence}; for example,
gravity and a linear spring on a circular orbit, Appendix~\ref{app:information-support}).
If $D_\mu$ is small but positive, the two candidate equations are difficult to distinguish, but not
impossible. The regime fixes this small difference, and the number of samples needed to distinguish
the two equations grows as $\sigma_\varepsilon^2/D_\mu$, where $\sigma_\varepsilon^2$ is the
variance of the error in the estimated derivatives
(Appendix~\ref{app:information-support}).
Hence, $D_\mu$ depends on the regime and on the two candidate equations, not
on the discovery algorithm.

Equation discovery builds candidate equations from a fixed library of $p$ functions,
$\Phi(x)=(\phi_1(x),\dots,\phi_p(x))$. Then, an equation is $f(x)=\xi^\top\Phi(x)$,
where $\xi\in\mathbb R^p$ holds its coefficients. For two equations with coefficients
$\xi$ and $\tilde\xi$, Eq.~\eqref{eq:vectorfielddiscrepancy} becomes
\begin{equation}
D_\mu(f,\tilde f)=(\xi-\tilde\xi)^\top M\,(\xi-\tilde\xi),
\qquad
M=\int\Phi(x)\Phi(x)^\top\,d\mu(x).
\label{eq:momentmatrix}
\end{equation}
We call $M$ the moment matrix: each entry $M_{ij}$ is the average of $\phi_i\phi_j$
on the visited states. A change of coefficients that gives zero in
Eq.~\eqref{eq:momentmatrix} does not change the vector field in the visited states, so
the data cannot see it. The smallest eigenvalue $\lambda_{\min}(M)$ measures the
weakest such change: it is zero if some change is invisible, and small if some change
is hard to see.

In practice, the invariant measure is not known, and we have only a finite sampled
trajectory. We therefore replace the average over $\mu$ with the average over the
samples:
\begin{equation}
\widehat M_N=\frac{1}{N}\sum_{n=1}^{N}\Phi(x_n)\Phi(x_n)^\top.
\label{eq:empirical-moment}
\end{equation}
Birkhoff's ergodic theorem~\citep{birkhoff1931proof} states that, along almost every
trajectory, averages over time converge to averages over $\mu$. The estimate
$\widehat M_N$ therefore converges to $M$ as the trajectory grows
(Appendix~\ref{app:sindy}). This has an important consequence: $M$ belongs to the
regime and to the library, not to the dataset. More data improve the estimate and
help to resolve small eigenvalues. However, no amount of data from the same regime
can turn a zero eigenvalue into a positive one.

The two indicators of the previous subsection are built on this object. Each
algorithm must tell apart a set of functions, and its indicator uses the moment matrix
of that set. SINDy must separate every library function from the others, so its
indicator uses the moment matrix of the whole library. PySR must separate the true
terms from the wrong ones, so its indicator uses the moment matrix of the true terms
after removing what the wrong terms can reproduce, up to a constant;
$\sigma_{\mathrm{sep}}^2$ is the smallest eigenvalue of this matrix
(Appendix~\ref{app:pysr}). The object is the same in both cases, and only the set of
functions changes. The regime therefore acts on both algorithms in the same way:
through the moment matrix of the functions they must tell apart.

Figure~\ref{fig:fMM} in the appendix shows how the regime shapes $M$ for L84. At the
fixed point R0, every sample is in the same state, so $M$ has rank one and
$\lambda_{\min}(M)=0$. The low-amplitude limit cycle R1 has
$\lambda_{\min}(M)\approx6\times10^{-5}$, about 25 times below R2. R2 and the chaotic
regimes all lie between $1.6\times10^{-3}$ and $5\times10^{-3}$, within about a factor
of three of each other, and their order does not follow the intensity of chaos. This
matches the recovery results: R0 and R1 are the hardest regimes, while R2--R7 follow no
fixed order (Fig.~\ref{fig:f2}). The weakest direction of $M$ thus separates the hard
regimes from the others, but it does not rank chaos above a well-excited limit cycle.
Appendix~\ref{app:l96} reports the same analysis for L96. Earlier work links attractor
coverage to SINDy conditioning~\citep{kaheman2020sindypi,lemus2024burst,bao2025fisher}
and uses ergodicity in exact-recovery guarantees~\citep{tran2017exact}. Here,
$\lambda_{\min}(M)$ gives one computable number per regime, which also covers the
fixed-point and limit-cycle cases.

\subsection{Regime, data, noise, and priors jointly set discoverability}

\begin{figure}[t!]
\centering
\includegraphics[width=\linewidth]{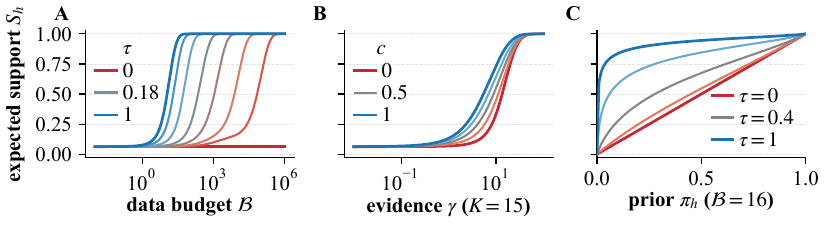}
\caption{\textbf{Data, excitation, and prior combine but do not replace each other.}
Expected posterior support $S_h$ of the true equation, in an abstract example with six
coefficients and fifteen competing equations. (A)~$S_h$ against the data budget
$\mathcal B$, for increasing excitation level $\tau$. (B)~$S_h$ against the evidence
$\gamma$ per competitor, for competitors that are more or less similar to each other
(correlation $c$); distinct competitors require more evidence. (C)~$S_h$ against the prior
probability $\pi_h$ of the true equation, for increasing excitation level $\tau$; without
excitation, $S_h$ equals $\pi_h$.
Details are in Appendix~\ref{app:information-support}.}
\label{fig:info-support}
\end{figure}

With finite,
noisy data, we must also ask how much evidence the data give for the true equation.
We measure it for an ideal observer, which chooses among the candidate equations using
their exact posterior probabilities~\citep{box1967discrimination,chaloner1995bayesian}.
No algorithm can do better on average. Let $h$ denote the true equation and $j$ be a
competitor, and let $D_\mu$ be their difference in the visited states
(Eq.~\eqref{eq:vectorfielddiscrepancy}). We call pairwise evidence the quantity
\begin{equation}
\gamma=\mathcal B\,D_\mu,
\qquad
\mathcal B=\frac{N_{\mathrm{eff}}}{\sigma_\varepsilon^2}.
\label{eq:pairwise-evidence}
\end{equation}
It is twice the Kullback--Leibler divergence between the data distributions of the two
equations: the larger $\gamma$, the less likely the observer confuses them. 
The data budget $\mathcal B$ grows with the effective number of samples $N_{\mathrm{eff}}$
(samples divided by the autocorrelation time) and shrinks with the derivative-error
variance $\sigma_\varepsilon^2$. The regime sets $D_\mu$. Halving $D_\mu$ therefore requires doubling $\mathcal B$.

With many competitors, pairwise evidence combines with the prior into one number:
the expected posterior support $S_h$, that is, the probability that the ideal
observer assigns to the true equation $h$, averaged over the noise
(Eq.~\eqref{eq:expected-posterior-support}). The prior is the probability $\pi_h$ that
the observer gives the true equation before seeing any data. We call excitation how
strongly the regime separates the candidates, as measured by $D_\mu$;
Figure~\ref{fig:info-support} varies it through a level $\tau\in[0,1]$, with $\tau=0$ giving
no separation. $S_h$ then depends
on three quantities: the data budget $\mathcal B$, the excitation, and the prior $\pi_h$.
Figure~\ref{fig:info-support} shows how they combine. More data and more excitation
reinforce each other (panel A). However, without excitation, no data budget helps:
$S_h$ stays equal to $\pi_h$ (panels A and C). A
favorable prior raises $S_h$, but it cannot create evidence that the data do not
contain (panel C).

Our experiments show each of these limits. Data volume and noise change the data
budget. R2 and the chaotic regimes reach near-perfect recovery once the budget is large
enough, and R1, with weak excitation, stays below them within the tested range. R0,
where every sample is in the same state, stays poorly recovered at every $N$
(Fig.~\ref{fig:f2}). Prior quality changes the prior. Even the oracle prior does not
rescue R0, and the null prior, which excludes the true terms, fails in every regime
despite clean data. For equation discovery, these results place a key limit outside the algorithm. The
dynamics fixes the distribution of states in the data, and this distribution decides
which equations the data can tell apart. Data volume, noise, and priors act within
this limit, but cannot remove it. The data alone cannot reveal what this distribution
does not show. Comparing discovery methods without controlling the dynamical regime
therefore mixes the difficulty of the data with the quality of the method.

\section{Discussion and Limitations}\label{sec:discussion}

Existing equation-discovery benchmarks compare different
systems~\citep{kaptanoglu2023benchmarking,gilpin2021chaos}. However, between systems,
the dynamical regime, the governing equations, and their algebraic complexity all
change at once. The difference in recovery can come either from the dynamics or
from the equation itself. Our design eliminates this confounder. We vary only the forcing
of the target dynamical system, so the governing equations keep the same form, and we evaluate every
regime with the same process. With the equations fixed, the regime alone strongly 
influences recovery. Therefore, benchmarks that mix systems mix the
difficulty of the data with the difficulty of the equation.

How recovery is scored also matters. Under the null prior, SINDy reaches a held-out
$R^2$ of $1.00$ with an equation that contains no true term. The circular-orbit
example shows that such an equation can even describe the wrong physics
(Appendix~\ref{app:information-support}). 
Predictive accuracy therefore cannot show that an equation was recovered. However, recent methods are still
evaluated this way. ODEFormer counts an equation as accurate when its simulated
trajectories reach $R^2$ above $0.9$~\citep{dascoli2024odeformer}, and LLM-SR
evaluates the discovered equations by their prediction error~\citep{shojaee2025llmsr}. However, the low
prediction error does not imply that the found equation is the true law~\citep{matsubara2022rethinking}.
Benchmarks should instead score the recovered structure directly. SRBench and
LLM-SRBench do so with symbolic checks~\citep{lacava2021contemporary,shojaee2025llmsrbench},
and Soft~F1 adds partial credit for coefficients. Predictive metrics should serve only
as a complement.

Even with the regime controlled and the structure scored, the two algorithms respond
to the data in different ways. SINDy uses the data through a ridge regression and a
hard threshold. Weak directions shrink its coefficients, and the threshold can then
remove true terms or keep the wrong ones. PySR instead builds expressions from the operator set, fits
their coefficients, and selects among them by accuracy and complexity. It fails when
wrong terms can imitate the true ones or when noise hides the part they cannot
imitate. The same data can therefore favor one algorithm over the other. This is
clearest under noise: the chaotic regime R7 tolerates noise better than R2 under
SINDy, but falls to the level of R1 under PySR with finite differences
(Fig.~\ref{fig:f2}).

Despite these differences, both algorithms read the same information. The discrepancy
$D_\mu$ (Eq.~\eqref{eq:vectorfielddiscrepancy}), the moment matrix
(Eq.~\eqref{eq:momentmatrix}), and the expected support $S_h$
(Fig.~\ref{fig:info-support}) describe what is in the data, independently of the
method. The smallest eigenvalue of the empirical moment matrix
(Eq.~\eqref{eq:empirical-moment}) can even be computed before any discovery and
without knowing the true equation, as a cheap check of whether the data can separate
the candidate functions.
This leads us to a broader conclusion: what
matters is whether the trajectory visits states where the candidate functions behave
differently, hence are distinguishable, and this is decided by the invariant 
measure of the regime. Therefore, regime labels are only rough proxies. 
A trajectory confined to a small region, such as a fixed
point or a small limit cycle, leaves some candidates hard or impossible to tell apart.
A trajectory that visits a wider and more varied set of states usually separates them
better, whether it is periodic or chaotic. The same holds for transient data, often 
discarded as burn-in: they help only when they excite the directions that the settled 
trajectory leaves weak (Proposition~\ref{prop:transient}).
How much variety the trajectory must show also depends on the amount of prior information. 
The more information in the prior, the fewer differences the
data need to reveal: with the oracle prior, even the weakly excited limit cycle R1 is
perfectly recovered (Fig.~\ref{fig:f2}D,H). When we consider a general candidate class, the
trajectory must provide more information, 
and chaos may be necessary~\citep{discoverability2025}. Chaos can therefore
help discovery, but neither a positive Lyapunov exponent nor stronger chaos guarantees
it.

These conclusions have several limitations. The experiments use deterministic
synthetic systems with full-state observation. Partial observation may require
delay-coordinate constructions~\citep{botvinick2024invariant} or latent-state models.
The SINDy indicator needs the true support to count true and false discoveries, so it
explains recovery rather than predicting it in advance. Without noise, it also
reproduces SINDy's first step, so part of its correlation with recovery reflects the
algorithm itself. The PySR indicator uses a combination chosen in Lorenz-84. It
transfers unchanged to the other systems, but it is not a probability model of
evolutionary search, and on Lorenz-84 its correlation weakens once the noise level is
controlled (Appendix~\ref{si:partial-correlation}). The eigenvalue check separates the regimes of our
systems, but we have not tested it as a stand-alone predictor of recovery. The
within-chaos analysis has only five chaotic regimes per system: enough to show that
excitation is not monotone in chaos intensity but not to estimate a general trend.
Finally, we study sparse regression on a fixed library and symbolic search. Whether
the same picture holds for neural discovery models remains open. Within these limits, 
our results make the dynamical regime a central variable of
symbolic equation discovery. It acts through the states the trajectory visits, and
this interacts with data, noise, priors, and the algorithm to decide whether an
equation can be recovered. Collecting data, developing methods, and comparing them
should, therefore, all account for the regime the data come from.

\section{Experimental Setup}\label{sec:experimental-setup}

\paragraph{Systems and regimes.}
We use Lorenz-84 (L84) as the controlled system. Varying only its forcing parameter
$F$ moves the dynamics across a fixed point, two limit-cycle branches, and five
chaotic regimes, while the functional form and the other parameters of the equations
stay fixed. We classify regimes from the Lyapunov spectrum. The second exponent
distinguishes limit cycles from quasiperiodic motion, and the attractor amplitude
separates the two limit-cycle branches. Lorenz-96 (L96) serves as a held-out system
with more dimensions, and polynomial, rational, and sinusoidal-rational jerk systems
test transfer beyond polynomial equations. Appendix~\ref{app:experimental-details}
gives the full regime-selection procedure and reference values.

\paragraph{Data and discovery algorithms.}
For each regime, we generate five independent, fully observed trajectories after a burn-in period.
Each contains a training pool of $10{,}000$ observations, whose first $N$ form a
training set of size $N$, and a held-out test block.
Noiseless experiments use derivatives computed from the governing equations. The noise
and prior-quality experiments use $N=5{,}000$. Noise is Gaussian and is added to each
state variable with a standard deviation equal to $\eta$ times that variable's standard
deviation. Derivatives are then estimated by second-order finite differences or by a
degree-3 Savitzky--Golay fit, which smooths the states and estimates the derivatives
together. The noise experiment uses four of the eight regimes, chosen before the
experiment to span the observed behaviors: both limit cycles (R1, R2) and the least and
most chaotic regimes (R3, R7). SINDy uses sequentially thresholded least squares on a
degree-3 polynomial library, with ridge parameter and threshold both fixed at $0.05$.
PySR uses the operators $\{+,-,\times\}$, a maximum complexity of 20, and its
\texttt{best} selection rule, and fits each trajectory from five independent
initializations. The prior levels change SINDy's library and PySR's operators and
input terms (Appendix~\ref{app:experimental-details}). The true equation of each
trajectory uses that trajectory's own forcing value, both for the derivatives and for
the scoring.

\paragraph{Evaluation and mechanistic quantities.}
Soft~F1 is computed for each state equation and then averaged over the equations of a
fit. The moment matrix and both indicators use the same trajectories, candidate terms,
and experimental conditions as the fits they describe
(Appendices~\ref{app:sindy} and~\ref{app:pysr}).

\paragraph{Replication and statistical analysis.}
Each SINDy cell contains five independent trajectory fits, and each PySR cell up to 25 fits, from five trajectories and five
initializations. Fig.~\ref{fig:f2} reports means and standard errors across fits. We
measure the link between each indicator and Soft~F1 by a Spearman correlation computed
for each state equation across experimental cells, then averaged over equations.
Confidence intervals come from a bootstrap that resamples whole trajectories, because
cells that share a trajectory are not independent. Appendix~\ref{app:experimental-details}
reports the integration settings, libraries, hyperparameters, noise propagation, and
uncertainty procedures.



\section*{Author contributions}
M.G.\ conceived the study, built the theoretical frameworks, ran all simulations and
analyses, and wrote the first draft of the manuscript. F.A.\ and P.L.\ supervised the
project, contributed to the theoretical development and interpretation of results, and
reviewed the manuscript. P.L.\ secured the funding that supported this research.

\section*{Competing interests}
The authors declare no competing interests.

\section*{Acknowledgments}
This work was supported by the European Union's Horizon Europe research and innovation
action (NECCTON, Grant Agreement No 101081273) and co-funded by the European Union
--- NextGenerationEU via the TeRABIT Project (IR0000022, PNRR Mission 4, Component 2,
Action 3.1).

\section*{AI use statement}
Generative AI tools were used to assist with editing manuscript text, writing
code, conducting literature searches and review, and checking mathematical
consistency and notation. All AI-assisted outputs were reviewed and verified
by the authors, who take responsibility for the final content of the paper.

\section*{Reproducibility statement}
Appendix~\ref{app:experimental-details} documents the experimental protocol,
including data generation, numerical integration, algorithmic settings, and
statistical analysis. The theoretical derivations and proofs are provided in
the supplementary appendices. The supplementary material contains the implementation, configuration
files, and instructions required to regenerate the synthetic data and rerun
the reported experiments.

\bibliographystyle{unsrtnat}
\bibliography{discov_skeleton}

\begin{thebibliography}{35}
\providecommand{\natexlab}[1]{#1}
\providecommand{\url}[1]{\texttt{#1}}
\expandafter\ifx\csname urlstyle\endcsname\relax
  \providecommand{\doi}[1]{doi: #1}\else
  \providecommand{\doi}{doi: \begingroup \urlstyle{rm}\Url}\fi

\bibitem[Brunton et~al.(2016)Brunton, Proctor, and
  Kutz]{brunton2016discovering}
Steven~L. Brunton, Joshua~L. Proctor, and J.~Nathan Kutz.
\newblock Discovering governing equations from data by sparse identification of
  nonlinear dynamical systems.
\newblock \emph{Proceedings of the National Academy of Sciences}, 113\penalty0
  (15):\penalty0 3932--3937, 2016.
\newblock \doi{10.1073/pnas.1517384113}.

\bibitem[Cranmer(2023)]{cranmer2023pysr}
Miles Cranmer.
\newblock Interpretable machine learning for science with {PySR} and
  {SymbolicRegression.jl}.
\newblock \emph{arXiv preprint arXiv:2305.01582}, 2023.

\bibitem[Schmidt and Lipson(2009)]{schmidt2009distilling}
Michael Schmidt and Hod Lipson.
\newblock Distilling free-form natural laws from experimental data.
\newblock \emph{Science}, 324\penalty0 (5923):\penalty0 81--85, 2009.

\bibitem[de~Fran{\c{c}}a et~al.(2025)de~Fran{\c{c}}a, Virgolin, Kommenda,
  Majumder, Cranmer, et~al.]{srbench2024}
Fabr{\'i}cio~Olivetti de~Fran{\c{c}}a, Marco Virgolin, Michael Kommenda, M.~S.
  Majumder, Miles Cranmer, et~al.
\newblock {SRBench++}: Principled benchmarking of symbolic regression with
  domain-expert interpretation.
\newblock \emph{IEEE Transactions on Evolutionary Computation}, 29\penalty0
  (4):\penalty0 1127--1137, 2025.
\newblock \doi{10.1109/TEVC.2024.3423681}.

\bibitem[Kaptanoglu et~al.(2023)Kaptanoglu, Zhang, Nicolaou, Fasel, and
  Brunton]{kaptanoglu2023benchmarking}
Alan~A. Kaptanoglu, Linan Zhang, Zachary~G. Nicolaou, Urban Fasel, and
  Steven~L. Brunton.
\newblock Benchmarking sparse system identification with low-dimensional chaos.
\newblock \emph{Nonlinear Dynamics}, 111:\penalty0 13143--13164, 2023.
\newblock \doi{10.1007/s11071-023-08525-4}.

\bibitem[Gilpin(2021)]{gilpin2021chaos}
William Gilpin.
\newblock Chaos as an interpretable benchmark for forecasting and data-driven
  modelling.
\newblock In Joaquin Vanschoren and Serena Yeung, editors, \emph{Proceedings of
  the Neural Information Processing Systems Track on Datasets and Benchmarks},
  volume~1. Curran Associates, Inc., 2021.

\bibitem[Shumaylov et~al.(2025)Shumaylov, Zaika, Scholl, Kutyniok, Horesh, and
  Sch{\"o}nlieb]{discoverability2025}
Zakhar Shumaylov, Peter Zaika, Philipp Scholl, Gitta Kutyniok, Lior Horesh, and
  Carola-Bibiane Sch{\"o}nlieb.
\newblock When is a system discoverable from data? {D}iscovery requires chaos.
\newblock \emph{arXiv preprint arXiv:2511.08860}, 2025.

\bibitem[Markovsky et~al.(2022)Markovsky, Prieto-Araujo, and
  D{\"o}rfler]{markovsky2022data}
Ivan Markovsky, Eduardo Prieto-Araujo, and Florian D{\"o}rfler.
\newblock On the persistency of excitation.
\newblock \emph{Automatica}, 147:\penalty0 110657, 2022.
\newblock \doi{10.1016/j.automatica.2022.110657}.

\bibitem[{\AA}str{\"o}m and Wittenmark(2008)]{astrom2008adaptive}
Karl~Johan {\AA}str{\"o}m and Bj{\"o}rn Wittenmark.
\newblock \emph{Adaptive Control}.
\newblock Dover Publications, 2nd edition, 2008.

\bibitem[Bao and Kutz(2025)]{bao2025fisher}
Yuxuan Bao and J.~Nathan Kutz.
\newblock Information theory and discriminative sampling for model discovery.
\newblock \emph{arXiv preprint arXiv:2512.16000}, 2025.
\newblock \doi{10.48550/arxiv.2512.16000}.

\bibitem[Tran and Ward(2017)]{tran2017exact}
Giang Tran and Rachel Ward.
\newblock Exact recovery of chaotic systems from highly corrupted data.
\newblock \emph{Multiscale Modeling \& Simulation}, 15\penalty0 (3):\penalty0
  1108--1129, 2017.
\newblock \doi{10.1137/16M1086637}.
\newblock arXiv:1607.01067.

\bibitem[Schaeffer et~al.(2018)Schaeffer, Tran, and Ward]{schaeffer2018extreme}
Hayden Schaeffer, Giang Tran, and Rachel Ward.
\newblock Extracting sparse high-dimensional dynamics from limited data.
\newblock \emph{SIAM Journal on Applied Mathematics}, 78\penalty0 (6):\penalty0
  3279--3295, 2018.
\newblock \doi{10.1137/17M1120792}.
\newblock arXiv:1707.08528.

\bibitem[Schaeffer et~al.(2020)Schaeffer, Tran, Ward, and
  Zhang]{schaeffer2020extracting}
Hayden Schaeffer, Giang Tran, Rachel Ward, and Linan Zhang.
\newblock Extracting structured dynamical systems using sparse optimization
  with very few samples.
\newblock \emph{Multiscale Modeling \& Simulation}, 18\penalty0 (4):\penalty0
  1435--1461, 2020.
\newblock \doi{10.1137/18M1194730}.
\newblock arXiv:1805.04158.

\bibitem[Ho et~al.(2020)Ho, Schaeffer, Tran, and Ward]{ho2018recovery}
Lam Si~Tung Ho, Hayden Schaeffer, Giang Tran, and Rachel Ward.
\newblock Recovery guarantees for polynomial coefficients from weakly dependent
  data with outliers.
\newblock \emph{Journal of Approximation Theory}, 259:\penalty0 105472, 2020.
\newblock \doi{10.1016/j.jat.2020.105472}.

\bibitem[Matsubara et~al.(2024)Matsubara, Chiba, Igarashi, and
  Ushiku]{matsubara2022rethinking}
Yoshitomo Matsubara, Naoya Chiba, Ryo Igarashi, and Yoshitaka Ushiku.
\newblock Rethinking symbolic regression datasets and benchmarks for scientific
  discovery.
\newblock \emph{Journal of Data-centric Machine Learning Research}, 1\penalty0
  (3):\penalty0 1--38, 2024.
\newblock URL \url{https://openreview.net/forum?id=qrUdrXsiXX}.

\bibitem[Lorenz(1984)]{lorenz1984irregularity}
Edward~N. Lorenz.
\newblock Irregularity: {A} fundamental property of the atmosphere.
\newblock \emph{Tellus A}, 36\penalty0 (2):\penalty0 98--110, 1984.

\bibitem[Lorenz(1996)]{lorenz1996predictability}
Edward~N. Lorenz.
\newblock Predictability: {A} problem partly solved.
\newblock In \emph{Proc.\ ECMWF Seminar on Predictability, Vol.\ 1}, pages
  1--18. ECMWF, 1996.

\bibitem[Sprott(1997)]{sprott1997some}
Julien~Clinton Sprott.
\newblock Some simple chaotic jerk functions.
\newblock \emph{American Journal of Physics}, 65\penalty0 (6):\penalty0
  537--543, 1997.
\newblock \doi{10.1119/1.18585}.

\bibitem[Broyden(1970)]{broyden1970convergence}
C.~G. Broyden.
\newblock The convergence of a class of double-rank minimization algorithms 1.
  general considerations.
\newblock \emph{IMA Journal of Applied Mathematics}, 6\penalty0 (1):\penalty0
  76--90, 1970.
\newblock \doi{10.1093/imamat/6.1.76}.

\bibitem[Delahunt and Kutz(2022)]{delahunt2022toolkit}
Charles~B. Delahunt and J.~Nathan Kutz.
\newblock A toolkit for data-driven discovery of governing equations in
  high-noise regimes.
\newblock \emph{IEEE Access}, 10:\penalty0 31210--31234, 2022.
\newblock \doi{10.1109/ACCESS.2022.3159335}.

\bibitem[Cortiella et~al.(2021)Cortiella, Park, and
  Doostan]{cortiella2020sparse}
Alexandre Cortiella, Kwang-Chun Park, and Alireza Doostan.
\newblock Sparse identification of nonlinear dynamical systems via reweighted
  $\ell_1$-regularized least squares.
\newblock \emph{Computer Methods in Applied Mechanics and Engineering},
  376:\penalty0 113620, 2021.
\newblock \doi{10.1016/j.cma.2020.113620}.

\bibitem[dos Reis et~al.(2024)dos Reis, Caminha, and
  Penna]{reis2024benchmarking}
L.~G.~A. dos Reis, V.~L. P.~S. Caminha, and T.~J.~P. Penna.
\newblock Benchmarking symbolic regression constant optimization schemes.
\newblock \emph{arXiv preprint arXiv:2412.02126}, 2024.
\newblock \doi{10.48550/arxiv.2412.02126}.

\bibitem[Birkhoff(1931)]{birkhoff1931proof}
George~D. Birkhoff.
\newblock Proof of the ergodic theorem.
\newblock \emph{Proceedings of the National Academy of Sciences}, 17\penalty0
  (12):\penalty0 656--660, 1931.

\bibitem[Kaheman et~al.(2020)Kaheman, Kutz, and Brunton]{kaheman2020sindypi}
Kadierdan Kaheman, J.~Nathan Kutz, and Steven~L. Brunton.
\newblock {SINDy-PI}: a robust algorithm for parallel implicit sparse
  identification of nonlinear dynamics.
\newblock \emph{Proceedings of the Royal Society A}, 476\penalty0
  (2242):\penalty0 20200279, 2020.
\newblock \doi{10.1098/rspa.2020.0279}.

\bibitem[Lemus and Herrmann(2024)]{lemus2024burst}
Jos{\'e}~Antonio Lemus and Bj{\"o}rn Herrmann.
\newblock Multi-objective {SINDy} for parameterized model discovery from single
  transient trajectory data.
\newblock \emph{Nonlinear Dynamics}, 113:\penalty0 10911--10927, 2024.
\newblock \doi{10.1007/s11071-024-10825-2}.

\bibitem[Box and Hill(1967)]{box1967discrimination}
George E.~P. Box and W.~J. Hill.
\newblock Discrimination among mechanistic models.
\newblock \emph{Technometrics}, 9\penalty0 (1):\penalty0 57--71, 1967.
\newblock \doi{10.1080/00401706.1967.10490441}.

\bibitem[Chaloner and Verdinelli(1995)]{chaloner1995bayesian}
Kathryn Chaloner and Isabella Verdinelli.
\newblock Bayesian experimental design: A review.
\newblock \emph{Statistical Science}, 10\penalty0 (3):\penalty0 273--304, 1995.
\newblock \doi{10.1214/ss/1177009939}.

\bibitem[d'Ascoli et~al.(2024)d'Ascoli, Becker, Mathis, Schwaller, and
  Kilbertus]{dascoli2024odeformer}
St{\'e}phane d'Ascoli, S{\"o}ren Becker, Alexander Mathis, Philippe Schwaller,
  and Niki Kilbertus.
\newblock {ODEFormer}: Symbolic regression of dynamical systems with
  transformers.
\newblock In \emph{International Conference on Learning Representations}, 2024.

\bibitem[Shojaee et~al.(2025{\natexlab{a}})Shojaee, Meidani, Gupta,
  Barati~Farimani, and Reddy]{shojaee2025llmsr}
Parshin Shojaee, Kazem Meidani, Shashank Gupta, Amir Barati~Farimani, and
  Chandan~K. Reddy.
\newblock {LLM-SR}: Scientific equation discovery via programming with large
  language models.
\newblock In \emph{International Conference on Learning Representations},
  2025{\natexlab{a}}.

\bibitem[La~Cava et~al.(2021)La~Cava, Orzechowski, Burlacu, de~Fran{\c{c}}a,
  Virgolin, Jin, Kommenda, and Moore]{lacava2021contemporary}
William La~Cava, Patryk Orzechowski, Bogdan Burlacu, Fabr{\'i}cio~Olivetti
  de~Fran{\c{c}}a, Marco Virgolin, Ying Jin, Michael Kommenda, and Jason~H.
  Moore.
\newblock Contemporary symbolic regression methods and their relative
  performance.
\newblock In \emph{Proceedings of the Neural Information Processing Systems
  Track on Datasets and Benchmarks}, 2021.

\bibitem[Shojaee et~al.(2025{\natexlab{b}})Shojaee, Nguyen, Meidani,
  Barati~Farimani, Doan, and Reddy]{shojaee2025llmsrbench}
Parshin Shojaee, Ngoc-Hieu Nguyen, Kazem Meidani, Amir Barati~Farimani, Khoa~D.
  Doan, and Chandan~K. Reddy.
\newblock {LLM-SRBench}: A new benchmark for scientific equation discovery with
  large language models.
\newblock In \emph{International Conference on Machine Learning},
  2025{\natexlab{b}}.

\bibitem[Botvinick-Greenhouse et~al.(2025)Botvinick-Greenhouse, Martin, and
  Yang]{botvinick2024invariant}
Jonah Botvinick-Greenhouse, Robert Martin, and Yunan Yang.
\newblock Invariant measures in time-delay coordinates for unique dynamical
  system identification.
\newblock \emph{Physical Review Letters}, 135\penalty0 (16):\penalty0 167202,
  2025.
\newblock \doi{10.1103/ppys-lx68}.

\bibitem[Donoho and Elad(2003)]{donoho2003optimally}
David~L. Donoho and Michael Elad.
\newblock Optimally sparse representation in general (nonorthogonal)
  dictionaries via $\ell^1$ minimization.
\newblock \emph{Proceedings of the National Academy of Sciences}, 100\penalty0
  (5):\penalty0 2197--2202, 2003.
\newblock \doi{10.1073/pnas.0437847100}.

\bibitem[Benettin et~al.(1980)Benettin, Galgani, Giorgilli, and
  Strelcyn]{benettin1980lyapunov}
Giancarlo Benettin, Luigi Galgani, Antonio Giorgilli, and Jean-Marie Strelcyn.
\newblock Lyapunov characteristic exponents for smooth dynamical systems and
  for {Hamiltonian} systems; a method for computing all of them. {Part 1}:
  Theory.
\newblock \emph{Meccanica}, 15\penalty0 (1):\penalty0 9--20, 1980.
\newblock \doi{10.1007/BF02128236}.

\bibitem[Kaptanoglu et~al.(2022)Kaptanoglu, de~Silva, Fasel, Kaheman,
  Goldschmidt, Callaham, Delahunt, Zheng, Mann, Kutz, and
  Brunton]{kaptanoglu2022pysindy}
Alan~A. Kaptanoglu, Brian~M. de~Silva, Urban Fasel, Kadierdan Kaheman, Andy~J.
  Goldschmidt, Jared~L. Callaham, Charles~B. Delahunt, Zachary~G. Zheng, Joshua
  Mann, J.~Nathan Kutz, and Steven~L. Brunton.
\newblock {PySINDy}: A comprehensive {Python} package for robust sparse system
  identification.
\newblock \emph{Journal of Open Source Software}, 7\penalty0 (69):\penalty0
  3994, 2022.
\newblock \doi{10.21105/joss.03994}.

\end{thebibliography}

\clearpage
\appendix
\counterwithin{figure}{section}
\counterwithin{table}{section}
\renewcommand{\theequation}{\thesection.\arabic{equation}}
\renewcommand{\thefigure}{\thesection.\arabic{figure}}
\renewcommand{\thetable}{\thesection.\arabic{table}}
\renewcommand{\theproposition}{\thesection.\arabic{proposition}}
\renewcommand{\theHfigure}{appendix.\thesection.\arabic{figure}}
\renewcommand{\theHtable}{appendix.\thesection.\arabic{table}}
\renewcommand{\theHequation}{appendix.\thesection.\arabic{equation}}

\section{Dynamical regime reference values}\label{app:regime-reference-values}

Table~\ref{tab:regime-reference-values} lists, for both systems, the reference forcing $F$ of each fixed-point
and chaotic regime and the attractor-amplitude range defining each data-driven
limit-cycle sub-regime (Appendix~\ref{app:experimental-details}, \textit{Numerical integration and regime selection}).
$F$ here is the regime reference value; the forcing of individual L84 trajectories can differ from it
(22 of 40 trajectories; Appendix~\ref{app:experimental-details}, \textit{Trajectory data and library construction}), and scoring uses the
forcing of the fitted trajectory. Values are taken from the per-regime classification records used
throughout the paper.

\begin{table}[h!]
\caption{Dynamical regime reference values for Lorenz-84 and Lorenz-96.}
\label{tab:regime-reference-values}
\centering
\begin{tabular}{llrrl}
\hline
System & Regime & $F$ & $\lambda_1$ & Notes \\
\hline
L84 & R0 (fp)    & $0.500$ & $-0.150$ & \\
L84 & R1 (lc)    & $4.798$ (ref.) & $0.000$ & $\sigma_x\in[0,\,0.567]$ \\
L84 & R2 (lc)    & $5.727$ (ref.) & $0.000$ & $\sigma_x\geq 0.590$ \\
L84 & R3 (chaos) & $8.283$ & $0.073$ & \\
L84 & R4 (chaos) & $8.515$ & $0.142$ & \\
L84 & R5 (chaos) & $7.934$ & $0.168$ & \\
L84 & R6 (chaos) & $8.747$ & $0.189$ & \\
L84 & R7 (chaos) & $7.586$ & $0.207$ & \\
\hline
L96 & R0 (fp)    & $0.200$ & $-0.300$ & \\
L96 & R1 (lc)    & $0.970$ (ref.) & $0.000$ & $\sigma_x\in[0,\,0.253]$ \\
L96 & R2 (lc)    & $3.879$ (ref.) & $0.000$ & $\sigma_x\geq 0.375$ \\
L96 & R3 (chaos) & $8.364$ & $0.507$ & \\
L96 & R4 (chaos) & $9.212$ & $0.754$ & \\
L96 & R5 (chaos) & $10.061$ & $0.932$ & \\
L96 & R6 (chaos) & $10.788$ & $1.133$ & \\
L96 & R7 (chaos) & $11.636$ & $1.348$ & \\
\hline
\end{tabular}
\end{table}

Limit-cycle regimes are data-driven amplitude bins rather than single pinned $F$
values (Appendix~\ref{app:experimental-details}): the $F$ entry shown is the single reference trajectory's forcing
value, but any $F$ within the regime's classification band (a range of 6--24 grid
points depending on regime, from the 100-point sweep) produces the same LC
classification and is pooled into the same $\sigma_x$ bin. Chaotic-regime $F$ values
are not monotonic in regime index for L84 (R3--R7 are ordered by increasing
$\lambda_1$, not by $F$, since $\lambda_1(F)$ is non-monotonic through the L84
period-doubling cascade); L96's chaotic $F$ values are monotonic in regime index
because $\lambda_1(F)$ increases monotonically over the sampled range.

\section{SINDy mechanistic model: full derivation}\label{app:sindy}

\subsection{Exact first-step ridge--threshold model}

STLSQ alternates ridge fits and hard thresholding.  We first isolate the one step
that can be characterized without assumptions about every later active-set refit.
For a realized (possibly noisy) library \(\widetilde\Theta\), ridge parameter
\(\alpha\), threshold \(\lambda\), and response
\[
\widetilde y=\widetilde\Theta\xi_{\mathrm{gt}}+\varepsilon_{\mathrm{eff}},
\qquad
\varepsilon_{\mathrm{eff}}
=\varepsilon_{\dot x}-\Delta\Theta\xi_{\mathrm{gt}},
\]
define
\[
H_\alpha=\widetilde\Theta^\top\widetilde\Theta+\alpha I,
\quad
R_\alpha=H_\alpha^{-1}\widetilde\Theta^\top\widetilde\Theta,
\quad
A_\alpha=H_\alpha^{-1}\widetilde\Theta^\top.
\]
The first ridge estimate obeys the exact decomposition
\begin{equation}
\widehat\xi=R_\alpha\xi_{\mathrm{gt}}+A_\alpha\varepsilon_{\mathrm{eff}}.
\label{eq:sindy-exact-ridge}
\end{equation}
The first term includes deterministic shrinkage and feature mixing; the second includes
response noise and errors in variables (EIV).  If
\(\widetilde\Theta=U\Sigma V^\top\), a singular direction with singular value
\(\sigma_k\) has signal shrinkage and response-noise gain
\[
w_k=\frac{\sigma_k^2}{\sigma_k^2+\alpha},
\qquad
g_k=\frac{\sigma_k}{\sigma_k^2+\alpha}.
\]
These are different functions.  In particular, for \(\alpha>0\), \(g_k\) is largest
at \(\sigma_k=\sqrt\alpha\), not necessarily at the smallest singular value.  Moreover,
thresholding occurs in the feature basis, whereas \(w_k\) is diagonal in the right-singular
basis.  A formula based only on \(w_{\min}\) is therefore an exact survival condition for
a feature only under an explicit alignment assumption; otherwise it is a worst-direction
proxy.  Equation~\eqref{eq:sindy-exact-ridge} also displays a noiseless false-positive
mechanism: for a wrong feature \(j\), \(e_j^\top R_\alpha\xi_{\mathrm{gt}}\) need not vanish.

For the conditional calculation used by the indicator, assume
\[
\varepsilon_{\mathrm{eff}}\mid\widetilde\Theta
\sim\mathcal N(b_{\mathrm{eff}},\Omega_{\mathrm{eff}}).
\]
Then
\[
\mu_\xi=R_\alpha\xi_{\mathrm{gt}}+A_\alpha b_{\mathrm{eff}},
\qquad
\Sigma_\xi=A_\alpha\Omega_{\mathrm{eff}}A_\alpha^\top.
\]
Let \(\mu_j=(\mu_\xi)_j\) and \(s_j^2=(\Sigma_\xi)_{jj}\).  The marginal probability
that coordinate \(j\) falls inside the threshold is
\begin{equation}
p_{\mathrm{in},j}
=\Phi\!\left(\frac{\lambda-\mu_j}{s_j}\right)
-\Phi\!\left(\frac{-\lambda-\mu_j}{s_j}\right),
\qquad p_{\mathrm{out},j}=1-p_{\mathrm{in},j},
\label{eq:sindy-threshold-prob}
\end{equation}
with the zero-variance limit interpreted deterministically.  No independence between
coefficient estimates is required because only marginal probabilities enter the expected
counts.  If \(S_{\mathrm{gt}}\) is the true support and a true term absent from the candidate
library is counted as a certain false negative, define
\[
\mathbb E[\mathrm{FN}]=n_{\mathrm{missing}}
+\sum_{j\in S_{\mathrm{gt}}\cap S_{\mathrm{lib}}}p_{\mathrm{in},j},
\quad
\mathbb E[\mathrm{FP}]=\sum_{j\notin S_{\mathrm{gt}}}p_{\mathrm{out},j},
\quad
\mathbb E[\mathrm{TP}]=|S_{\mathrm{gt}}|-\mathbb E[\mathrm{FN}].
\]
Our unified SINDy indicator is the F1 of these expected threshold counts,
\begin{equation}
\mathcal I_{\mathrm{SINDy}}
=\frac{2\mathbb E[\mathrm{TP}]}
{2\mathbb E[\mathrm{TP}]+\mathbb E[\mathrm{FP}]+\mathbb E[\mathrm{FN}]}.
\label{eq:fsindy-si}
\end{equation}
It is not generally \(\mathbb E[\mathrm{F1}]\), and it is not the probability of exact
final STLSQ recovery.  In the noiseless zero-covariance limit it reduces to the support
F1 after the first ridge--threshold step.  A deletion at this step is irreversible for
the monotone-support implementation used here; survival does not by itself prove final
multistep recovery.

For the noisy experiments, \(b_{\mathrm{eff}}\) and \(\Omega_{\mathrm{eff}}\) are
estimated by propagating
the specified Gaussian state-noise process through the actual finite-difference or
Savitzky--Golay derivative pipeline and rebuilding the noisy polynomial library.  Antithetic
Monte Carlo pairs estimate each coefficient's mean and variance, after which
Eq.~\eqref{eq:sindy-threshold-prob} is applied.  This captures EIV mean shifts, filtering,
temporal dependence, and design--response coupling at the stated moment-matched marginal
Gaussian level. We use 32 antithetic replicates. The correlations with Soft~F1 in the noise
experiment do not depend on this budget: 8, 16, and 32 replicates give \(0.914\), \(0.915\),
and \(0.917\) for L84 and \(0.888\), \(0.896\), and \(0.896\) for L96. The computation is
implemented in \texttt{analysis/validate\_sindy\_threshold\_indicator.py} of the supplementary code, with deterministic
seeds.

\subsection{Convergence proof: Birkhoff and Weyl}

Let $\Phi(x)\in\mathbb{R}^p$ be a fixed finite dictionary and let
$\Theta_N\in\mathbb{R}^{N\times p}$ have rows $\Phi(x_n)^\top$, where
$x_n=\mathcal{T}^n x_0$ and $\mathcal{T}=\varphi_{\Delta t}$ is the fixed-time
sampling map. Assume that $\mathcal{T}$ preserves an ergodic invariant probability
measure $\mu$, that $x_0$ is typical for the finite collection
$\{\phi_i\phi_j\}_{i,j=1}^p$, and that every product $\phi_i\phi_j$ belongs to
$L^1(\mu)$. Continuity of the dictionary on compact support is one sufficient way to
ensure the last condition, but is not necessary. Define
\[
  \widehat M_N=\frac{1}{N}\Theta_N^\top\Theta_N,
  \qquad
  M=\int \Phi(x)\Phi(x)^\top\,d\mu(x).
\]
If different trajectories lie on different ergodic components, the following
statement is applied separately to each component and its corresponding measure.

For every pair $(i,j)$, the Birkhoff ergodic theorem gives
\[
  (\widehat M_N)_{ij}
  =\frac{1}{N}\sum_{n=0}^{N-1}\phi_i(x_n)\phi_j(x_n)
  \longrightarrow
  \int\phi_i(x)\phi_j(x)\,d\mu(x)=M_{ij}
\]
for such a typical initial condition. Writing $D_N=\widehat M_N-M$, the fixed
finite dimension is the step that turns these scalar limits into a matrix limit:
\[
  \|D_N\|_F^2=\sum_{i,j=1}^p |(D_N)_{ij}|^2\longrightarrow0,
  \qquad
  \|D_N\|_2\leq\|D_N\|_F\longrightarrow0.
\]
This is an almost-everywhere infinite-time statement; it supplies neither a finite
burn-in certificate nor a finite-sample convergence rate.

The singular values $\sigma_k(\Theta_N)$ are the square roots of the eigenvalues of
$\Theta_N^\top\Theta_N$, with coefficient-space zero directions included when the
finite design is underdetermined. Hence the eigenvalues of $\widehat M_N$ are exactly
$\sigma_k(\Theta_N)^2/N$.
Weyl's inequality for real symmetric matrices,
\[
  |\lambda_k(A)-\lambda_k(B)| \;\leq\; \|A-B\|_2,
\]
applied with $A=\widehat M_N$ and $B=M$, gives
\[
  \left|\frac{\sigma_k(\Theta_N)^2}{N}-\lambda_k(M)\right|
  \leq \|\widehat M_N-M\|_2\longrightarrow0.
\]
If $\lambda_k(M)>0$, continuity of the square root yields
\[
  \frac{\sigma_k(\Theta_N)}{\sqrt N}
  \longrightarrow \sqrt{\lambda_k(M)},
  \qquad\text{equivalently}\qquad
  \sigma_k(\Theta_N)\sim\sqrt{N\lambda_k(M)}.
\]
If $\lambda_k(M)=0$, the valid general conclusion is instead
$\sigma_k(\Theta_N)=o(\sqrt N)$; a relative asymptotic equivalence to zero is not
defined. In particular, when $\lambda_{\min}(M)>0$,
\begin{equation}
  \frac{\sigma_{\min}(\Theta_N)}{\sqrt N}
  \longrightarrow\sqrt{\lambda_{\min}(M)}.
  \label{eq:sigmamin}
\end{equation}

\section{PySR mechanistic model: full derivation}\label{app:pysr}

\subsection{Moment matrix convergence for PySR}

The Birkhoff argument connects the structural channel of
$\mathcal{F}_{\mathrm{PySR}}$ to $\mu$ through the overcomplete population moment
matrix
\[
M_{\mathrm{full}}
=\int\Phi_{\mathrm{full}}(x)\Phi_{\mathrm{full}}(x)^\top\,d\mu(x).
\]
Partition the selected candidate terms into the true nonconstant block $\mathrm{NC}$,
the true constant block $\mathrm C$, and the wrong-candidate block $\mathrm W$. The
baseline/comparison block is $\mathrm B=\mathrm C\cup\mathrm W$: it is the space used
to reproduce the $\mathrm{NC}$ block, not a claim that all of its terms are always
present in every PySR expression.

Let $\theta_{i,N}$ be the sampled column for term $i$ and
$z_{i,N}=\theta_{i,N}/\|\theta_{i,N}\|_2$. We assume that each normalized term has
positive population second moment $M_{ii}=\int\phi_i^2\,d\mu>0$; a term that is zero
$\mu$-almost everywhere must be removed or treated by a separate numerical
convention. With $D=\operatorname{diag}(M_{\mathrm{full}})$, the entrywise Birkhoff
argument from Appendix~\ref{app:sindy} gives
\[
Z_N^\top Z_N\longrightarrow
\overline M_{\mathrm{full}}
=D^{-1/2}M_{\mathrm{full}}D^{-1/2},
\]
because
\[
z_{i,N}^\top z_{j,N}
=\frac{(\widehat M_N)_{ij}}
{\sqrt{(\widehat M_N)_{ii}(\widehat M_N)_{jj}}}.
\]
There is no additional $1/N$ factor: every column of $Z_N$ already has unit
Euclidean norm. In particular, the normalized constant column is
$\mathbf 1_N/\sqrt N$.

Write $Z_N=[Z_{\mathrm{NC},N},Z_{\mathrm B,N}]$ and define
\[
P_{\mathrm B,N}=Z_{\mathrm B,N}Z_{\mathrm B,N}^{\dagger},
\qquad
R_N=(I_N-P_{\mathrm B,N})Z_{\mathrm{NC},N}.
\]
Here $P_{\mathrm B,N}$ is the orthogonal projector onto
$\operatorname{col}(Z_{\mathrm B,N})$. Therefore, for every coefficient direction
$v$,
\[
\min_b\|Z_{\mathrm{NC},N}v-Z_{\mathrm B,N}b\|_2^2
=v^\top R_N^\top R_Nv.
\]
If the normalized empirical Gram matrix is partitioned as
\[
Z_N^\top Z_N=
\begin{pmatrix}
A_N&K_N\\ K_N^\top&H_N
\end{pmatrix},
\qquad H_N=Z_{\mathrm B,N}^\top Z_{\mathrm B,N},
\]
then the finite-sample projection identity is
\[
R_N^\top R_N=A_N-K_NH_N^\dagger K_N^\top.
\]
Thus $R_N^\top R_N$ is a residual Gram matrix, equivalently an empirical
generalized Schur complement; it is not in general a correlation matrix, because its
diagonal entries need not equal one.

Let $A$, $K$, and $H$ be the corresponding blocks of
$\overline M_{\mathrm{full}}$. Birkhoff and the fixed-dimensional norm argument give
$A_N\to A$, $K_N\to K$, and $H_N\to H$. If $H$ is positive definite, Weyl's
inequality makes $H_N$ invertible for all sufficiently large $N$, and inverse
continuity gives $H_N^{-1}\to H^{-1}$. If $H$ is singular, convergence alone is not
enough because the pseudoinverse is discontinuous at a rank change. Let
$s=\operatorname{rank}(H)$. Its $s$ positive eigenvalues imply
$\operatorname{rank}(H_N)\geq s$ eventually. Conversely, every population linear
relation among the baseline functions holds $\mu$-almost everywhere. Invariance of
$\mu$, followed by a countable union over sampled times and a finite union over a
basis of these relations, shows that all such relations hold along a typical sampled
orbit. Hence $\operatorname{rank}(H_N)\leq s$. The ranks therefore equal $s$ for all
sufficiently large $N$, so $H_N^\dagger\to H^\dagger$. This threshold is an
unspecified $N_0$, not generally $s$; no real-analyticity hypothesis is needed.

In both rank cases,
\[
R_N^\top R_N\longrightarrow
\overline M_{\mathrm{NC}\mid\mathrm B}
=A-KH^\dagger K^\top.
\]
Weyl's inequality and continuity of the square root then yield
\[
\sigma_{\min}(R_N)^2
\longrightarrow\lambda_{\min}(\overline M_{\mathrm{NC}\mid\mathrm B}),
\qquad
\sigma_{\min}(R_N)
\longrightarrow\sqrt{\lambda_{\min}(\overline M_{\mathrm{NC}\mid\mathrm B})}.
\]
The main text calls this quantity the separation, $\sigma_{\mathrm{sep}}:=\sigma_{\min}(R_N)$.
Again there is no $\sqrt N$ factor. A zero limiting eigenvalue means that at least one
nonzero true direction is exactly reproducible by the comparison block; it does not
mean that every true term is reproducible. A small positive value measures poor
separation for this dictionary and invariant measure, but is not by itself a complete
theorem of PySR recovery.

For the noise channel, the statistic is a nested least-squares
construction rather than the time average of one fixed observable. Fix one governing
equation, let $y_N$ be its sampled noiseless derivative response, and let
$\Psi_{-k,N}$ contain the remaining true nonconstant terms together with the true
constant. Define the minimum-norm first-stage fit and residual by
\[
\widehat c_{k,N}
\in\arg\min_c\frac1N\|y_N-\Psi_{-k,N}c\|_2^2,
\qquad
r_{k,N}=y_N-\Psi_{-k,N}\widehat c_{k,N}.
\]
For each wrong-candidate index $\ell\in\mathrm W$, fit this residual with that one
wrong column and an intercept:
\[
\varepsilon_{k,\ell,N}
=\min_{a,b}\frac1N
\|r_{k,N}-a\theta_{\ell,N}-b\mathbf1_N\|_2^2,
\qquad
\varepsilon_{\mathrm{wrong},k,N}
=\min_{\ell\in\mathrm W}\varepsilon_{k,\ell,N}.
\]

The corresponding population object must retain the first projection. Let
\[
c_k^\mu\in\arg\min_c
\int|f(x)-c^\top\Psi_{-k}(x)|^2\,d\mu(x),
\qquad
r_k(x)=f(x)-(c_k^\mu)^\top\Psi_{-k}(x),
\]
where the minimum-norm minimizer is selected when it is nonunique. Then, provided
the first- and second-stage population Gram blocks are invertible or their empirical
ranks are eventually stable as above,
\begin{equation}
\varepsilon_{\mathrm{wrong},k,N}
\longrightarrow
\min_{\ell\in\mathrm W}\min_{a,b}
\int|r_k(x)-a\phi_\ell(x)-b|^2\,d\mu(x).
\label{eq:epsilon-wrong-limit}
\end{equation}

The dependence of $r_{k,N}$ and of both fitted stages on $N$ is why Birkhoff cannot
be applied directly to the final displayed loss. Instead, Birkhoff is first applied
entrywise to the fixed feature products and response--feature products in the
first-stage normal equations. Inverse continuity, or pseudoinverse continuity under
eventual rank stability, then gives $\widehat c_{k,N}\to c_k^\mu$. If
$r_k(X_N)$ denotes the vector obtained by sampling the fixed population residual,
then
\[
\frac1N\|r_{k,N}-r_k(X_N)\|_2^2
=(\widehat c_{k,N}-c_k^\mu)^\top
\widehat M_N^{(-k)}(\widehat c_{k,N}-c_k^\mu)
\longrightarrow0.
\]
For each fixed $\ell$, Birkhoff applies to the products involving $r_k$, while
Cauchy--Schwarz controls the vanishing correction from replacing $r_k(X_N)$ by
$r_{k,N}$. The second-stage coefficients and minimized loss therefore converge.
Finally, the minimum can pass through the limit because the wrong-candidate set
$\mathrm W$ is fixed and finite.

In an exact noiseless model,
\[
r_k=\xi_k\,[\phi_k-\Pi_{-k}\phi_k],
\]
where $\Pi_{-k}$ is the $L^2(\mu)$ projection onto the remaining true terms and the
constant. Thus $\varepsilon_{\mathrm{wrong},k}$ is the portion of the population
leave-one-true-term-out residual contribution that cannot be reproduced by one wrong
candidate plus an intercept. It is not, in general, the raw distance from $\phi_k$ to
a rescaled wrong atom: the true coefficient and correlations among the true terms are
part of the object.

\emph{Noise channel.} For each true nonconstant term $k$ of an equation, the noise
factor compares $\varepsilon_{\mathrm{wrong},k}$ with the variance of the derivative
error, and the factors are multiplied:
\begin{equation}
Q_{\mathrm{noise}}
=\prod_{k\in\mathrm{NC}}
\min\!\left(1,\sqrt{\frac{\varepsilon_{\mathrm{wrong},k}}{\sigma_\varepsilon^2}}\right),
\qquad
\sigma_\varepsilon^2=\frac1N\sum_{n=1}^{N}\bigl(\widehat{\dot x}_n-\dot x_n-\overline e\bigr)^2 .
\label{eq:qnoise}
\end{equation}
Here $\widehat{\dot x}_n$ is the derivative of the equation estimated from the noisy
states by the derivative estimator of the experiment (finite differences or
Savitzky--Golay), $\dot x_n$ is the exact derivative, and $\overline e$ is the mean
error, so $\sigma_\varepsilon^2$ is the variance of the derivative error. Without noise,
$\sigma_\varepsilon^2=0$ and $Q_{\mathrm{noise}}=1$. Like the SINDy indicator,
$Q_{\mathrm{noise}}$ uses the true equation and the true derivatives, so it explains
recovery rather than predicting it without this knowledge.

\subsection{Functional-form selection: what is derived and what is selected}
\label{si:variant-sweep}

The residual geometry $\sigma_{\mathrm{sep}}$ and the population limits of
$\varepsilon_{\mathrm{wrong},k}$ follow from the projection and convergence results above.
The ratio $\sqrt{\varepsilon_{\mathrm{wrong},k}/\sigma_\varepsilon^2}$ is a dimensionless
amplitude-to-noise comparison. The cap at one, product over terms, quarter-power applied to
$\sigma_{\mathrm{sep}}$, and geometric-mean combination are modeling choices. In
particular, $Q_{\mathrm{noise}}$ is a soft conjunctive bottleneck, not a joint probability and
does not assume independence. The aggregation was selected by maximizing mean Spearman
correlation on L84 across all three experiments before L96 was consulted.

We report the full sensitivity grid: 5 monotone transforms of $\sigma_{\mathrm{sep}}$
(raw, $\sqrt{\cdot}$, $(\cdot)^{1/4}$, $(\cdot)^{1/8}$, $1-e^{-(\cdot)}$) $\times$ 4 combination
rules with $Q_{\mathrm{noise}}$ (product, arithmetic mean, geometric mean, harmonic mean),
each scored by the same per-dimension-averaged Spearman $\rho$ used throughout, averaged
over the three experiments. (A sixth transform, $\log(\sigma_{\mathrm{sep}}+\epsilon)$,
is excluded from the grid below because it can be negative, making the product/geometric/
harmonic combination rules ill-defined; it is only ever meaningfully paired with the
arithmetic mean, for the same reason.)

\begin{center}
\begin{tabular}{lrr}
\hline
 & L84 (selection) & L96 (held out) \\
\hline
Range of mean Spearman $\rho$ across the 20-member grid & $0.803$--$0.819$ & $0.840$--$0.870$ \\
Selected model ($(\cdot)^{1/4}$, geometric mean) & $0.819$ & $0.869$ \\
\hline
\end{tabular}
\end{center}

The trajectory-averaged plateau spans $\Delta\rho=0.016$ on L84 and
$\Delta\rho=0.030$ on L96: every
member of the grid supports the same qualitative conclusions this paper draws from
$\mathcal{F}_{\mathrm{PySR}}$. Because Spearman correlation is invariant to any monotone
rescaling of a single variable, much of this insensitivity is expected by construction; as a
statistic that is \emph{not} monotone-invariant, we also report the Pearson correlation
between $\log(\mathrm{score})$ and $\mathrm{logit}(\mathrm{Soft~F1})$, which is sensitive to
the specific functional form: $r=0.63$--$0.91$ across the six (system, experiment) cells (all
$p<10^{-100}$), confirming the relationship is not an artifact of rank-invariance alone.

\emph{Ablation.} Both factors vary only in the noise experiment; in the other two
experiments $Q_{\mathrm{noise}}=1$ and $\sigma_{\mathrm{sep}}$ is the whole score. In the
noise experiment, with correlations computed over individual runs (slot and
initialization) rather than slot means, the full score gives $\rho=0.759$ on L84 and
$0.845$ on L96. On L96, $\sigma_{\mathrm{sep}}^{1/4}$ alone gives $0.557$ and
$Q_{\mathrm{noise}}$ alone gives $0.786$, so both factors contribute. On L84,
$\sigma_{\mathrm{sep}}^{1/4}$ alone gives $0.270$ and $Q_{\mathrm{noise}}$ alone gives
$0.754$, so under noise the L84 correlation comes almost entirely from $Q_{\mathrm{noise}}$.
Averaging over the initializations of each slot before correlating, as in
Table~\ref{tab:noiseless}, gives the table values $0.81$ and $0.88$.

\section{Analytical theory: derivations and formal statements}\label{app:theory}

\begin{figure}[t!]
  \includegraphics[width=\linewidth]{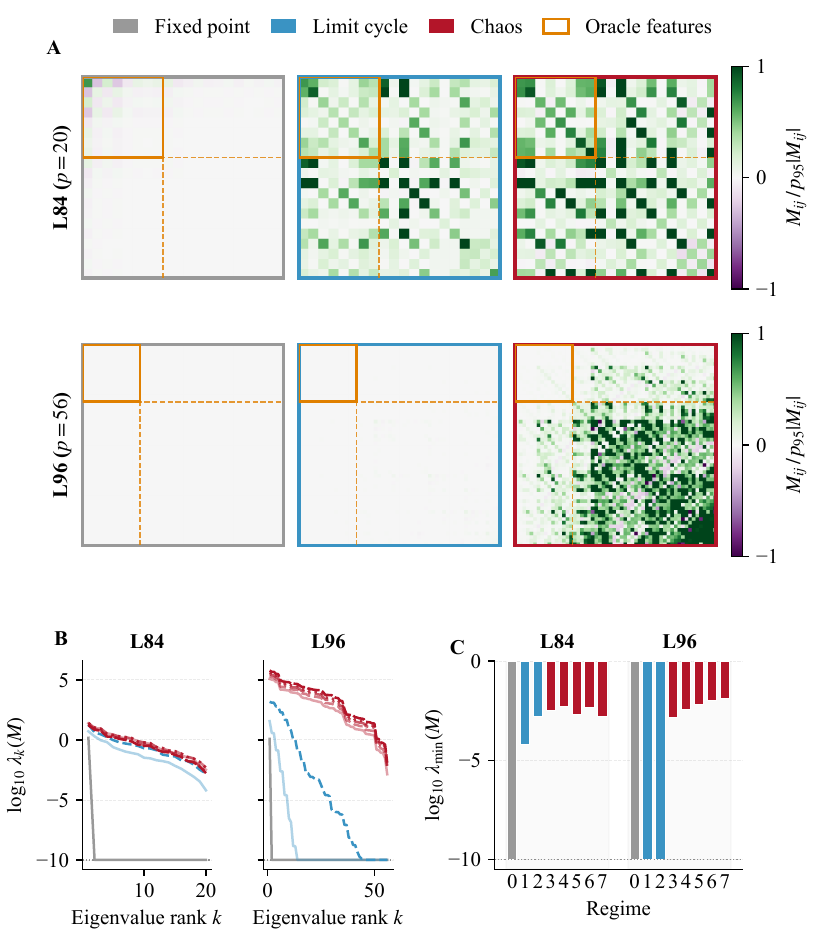}
  \caption{\textbf{The moment matrix encodes dynamical regime through its eigenvalue
    structure.}
    (\textit{A})~Raw moment matrices $M = (1/N)\,\Theta^\top\Theta$ for a
    representative fixed point (grey border), limit cycle (blue), and chaotic regime
    (red), shown for L84 ($p = 20$, top row) and L96 ($p = 56$, bottom row).
    The orange box highlights the subblock spanned by the ground-truth dictionary terms;
    dashed lines mark its boundary.
    At a fixed point all dictionary rows are identical, yielding a rank-1 matrix; the
    displayed limit cycles have weakly excited (L84) or exactly null (L96) directions; the displayed chaotic
    regimes excite all $p$ dictionary directions above the numerical floor.
    (\textit{B})~Eigenvalue spectra $\log_{10}(\lambda_k)$ vs.\ rank $k$ for every
    regime (L84 left, L96 right; line opacity increases with chaos intensity).
    Fixed-point regimes collapse to a single nonzero eigenvalue; chaotic regimes
    maintain all eigenvalues well above the numerical floor (dotted line).
    (\textit{C})~$\log_{10}\,\lambda_{\min}(M)$ per regime R$k$ (horizontal axis: $k$) for L84
    and L96; bars reaching the dotted numerical floor are zero to numerical precision.
    Across the sampled regimes, the spectral scale spans more than three orders of magnitude.
    In L84 it separates the hardest regimes (R0, R1) from the others, matching
    Fig.~\ref{fig:f2}, but it does not rank the chaotic regimes above the large limit
    cycle R2.}
  \label{fig:fMM}
\end{figure}

This appendix collects the formal results behind the moment-matrix analysis of the main
text. Figure~\ref{fig:fMM} shows how the regime shapes the moment matrix.
Appendix~\ref{si:exact-collinearity} shows that a zero direction of $M$ makes distinct
equations indistinguishable in any regime, and Appendix~\ref{si:spark} treats the fixed
point. Appendix~\ref{si:finite-time} bounds how fast a finite trajectory estimates $M$,
and Appendix~\ref{si:transient} states when transient data add information.

\subsection{Exact collinearity makes equations indistinguishable in any regime}
\label{si:exact-collinearity}

Let $D=\operatorname{diag}(M)$ have positive entries,
$\overline\Phi=D^{-1/2}\Phi$, and $\eta=D^{1/2}\xi$. This rescaling preserves
both predictions, $\Phi^\top\xi=\overline\Phi^\top\eta$, and coefficient
support. Let $\overline\Theta_N$ denote the corresponding sampled design,
whose $n$th row is $\overline\Phi(x_n)^\top$.

\medskip
\refstepcounter{proposition}\label{prop:equivalence}\noindent\textbf{Proposition~\theproposition} (On-support observational equivalence).
Assume the dictionary functions are continuous and that, for some $v\neq0$,
\[
v^\top\overline\Phi(x)=0\qquad\mu\text{-almost everywhere}.
\]
Then the relation holds at every $x\in\operatorname{supp}(\mu)$, and for every
$\alpha\in\mathbb R$,
\[
\overline\Phi(x)^\top(\eta^{\mathrm{gt}}+\alpha v)
=\overline\Phi(x)^\top\eta^{\mathrm{gt}}
\qquad\text{on }\operatorname{supp}(\mu).
\]
Thus the clean on-support data identify only the equivalence class
$\eta^{\mathrm{gt}}+\operatorname{span}\{v\}$, not a unique coefficient
vector. If the sampled map $\mathcal T$ preserves $\mu$, then for
$\mu$-almost every initial condition the same null relation holds at every
sampled time and hence
\[
\overline\Theta_Nv=0
\]
for every finite $N$.

\smallskip
\noindent\textit{Proof.}
The combination $\psi_v=v^\top\overline\Phi$ is continuous. If it were nonzero at
a point of $\operatorname{supp}(\mu)$, continuity would give a neighborhood on
which it is bounded away from zero; every neighborhood of a support point has
positive $\mu$-measure, contradicting $\psi_v=0$ almost everywhere. The coefficient
continuum follows by substitution. For the sampled statement, let
$E=\{x:\psi_v(x)\neq0\}$. Since $\mu(E)=0$ and $\mu$ is invariant,
$\mu(\mathcal T^{-n}E)=0$ for every $n$. The countable union over $n$ is still null,
so a typical orbit never meets $E$ at a sampled time. Ergodicity is not needed for
this last step.
\hfill$\square$

\smallskip
\refstepcounter{proposition}\label{cor:sparse-pairs}\noindent\textbf{Corollary~\theproposition} (Existence of sparse indistinguishable pairs).
Let $r=\|v\|_0$. Split $\operatorname{supp}(v)$ into disjoint sets $A$ and $B$
with cardinalities at most $\lceil r/2\rceil$, and define
$\zeta_1=v|_A$ and $\zeta_2=-v|_B$. Then
$\zeta_1-\zeta_2=v$, so the two distinct vectors generate identical sampled
predictions and each has sparsity at most $\lceil r/2\rceil$. This is an existence
result for some pair; it does not by itself show that the actual ground truth has an
equally sparse competitor.

\smallskip
\refstepcounter{proposition}\label{cor:target-sparsity}\noindent\textbf{Corollary~\theproposition} (Target-specific sparsity failure).
Let $I=\operatorname{supp}(\eta^{\mathrm{gt}})$. Suppose
\[
\operatorname{supp}(v)\cap I\neq\varnothing,
\qquad
|\operatorname{supp}(v)\setminus I|\leq1.
\]
Choose $j\in I$ with $v_j\neq0$, set
$\alpha=-\eta_j^{\mathrm{gt}}/v_j$, and define
$\widetilde\eta=\eta^{\mathrm{gt}}+\alpha v$. The $j$th active coordinate is
cancelled, existing active coordinates cannot enlarge the support, and at most one new
coordinate is activated. Therefore
\[
\|\widetilde\eta\|_0\leq\|\eta^{\mathrm{gt}}\|_0,
\]
while Proposition~\ref{prop:equivalence} gives identical clean predictions. The inequality is a
conclusion, not an assumption. Pairwise duplication of one active true term by one
inactive wrong term is the simplest special case. A singular generalized Schur
complement supplies a null direction involving the true and comparison blocks, but
does not by itself bound the number of newly activated wrong coordinates; it therefore
does not alone imply this target-specific sparse conclusion.

\smallskip
\noindent\textbf{Remark 1 (measurement-error experiment).}
Suppose the ground-truth and alternative vector fields agree on
$\operatorname{supp}(\mu)$, both ODEs have unique solutions near that support, and
they start from the same support point. The common clean trajectory solves both ODEs.
Applying the same post-integration observation channel
$Y=\mathcal O(X,\varepsilon)$ therefore gives the same observed-data law under both
models, and applying the same deterministic derivative estimator preserves that
equality. No data-only estimator can distinguish the two models within this
experiment and admitted model class. This covers the paper's measurement-noise
protocol, but not process noise that moves the latent path off the support.

Noisy covariates can nevertheless lift the numerical null direction of the
normalized design:
\[
v^\top\overline\Phi(x+\varepsilon)
=
\nabla\!\left(v^\top\overline\Phi\right)(x)^\top\varepsilon
+O(\|\varepsilon\|^2).
\]
Thus a noisy design may be full rank even though the latent equation remains
nonidentifiable; its unique regression solution can be a noise-dependent
pseudo-solution.

\smallskip
\noindent\textbf{Remark 2 (support, analyticity, and geometry).}
Continuity is sufficient for an almost-everywhere null relation to hold everywhere on
$\operatorname{supp}(\mu)$. Analyticity is needed only for stronger extension beyond
the support, and then only with an ambient open-set or absolute-continuity condition.
A dissipative chaotic SRB measure on a fractal attractor need not be absolutely
continuous with respect to ambient Lebesgue measure. Likewise, lower-dimensional
support can create dictionary-specific relations but does not force a chosen finite
dictionary to contain one; a literal redundancy such as $\phi_2=2\phi_1$ persists in
every regime.

The relevant object is therefore excitation of the chosen dictionary under $\mu$,
not attractor dimension or the label ``chaotic'' by itself. For example,
$\Phi=(1,x^2,y^2)^\top$ is singular on the unit circle because
$-1+x^2+y^2=0$, but is positive definite under the uniform measure on
$[-1,1]^2$. Conversely, the unit circle is perfectly conditioned for
$\Phi=(x,y)^\top$, whereas an invertible coordinate transform of a chaotic measure
can make two standardized features arbitrarily correlated and drive the normalized
minimum eigenvalue arbitrarily close to zero. Chaos can improve dictionary excitation
in a particular system, but supplies no universal spectral ordering.

\smallskip
\noindent\textbf{Remark 3 (comparison across regimes).}
Let $M_1$ and $M_2$ be the moment matrices of the same dictionary under two regimes.
By the Rayleigh characterization,
$\lambda_{\min}(M_r)=\min_{\|v\|_2=1}\int(v^\top\Phi)^2\,d\mu_r$: the smallest
eigenvalue is the smallest mean-square value of a unit combination of dictionary terms.
If $M_2-M_1\succeq0$, every combination gains energy and
$\lambda_{\min}(M_2)\geq\lambda_{\min}(M_1)$, but no chaos indicator, including a
positive Lyapunov exponent, implies this ordering. If $u_1$ is a unit eigenvector of
$\lambda_{\min}(M_1)$, then
\[
\lambda_{\min}(M_2)\leq u_1^\top M_2u_1
=\lambda_{\min}(M_1)+u_1^\top(M_2-M_1)u_1,
\]
so conditioning degrades whenever the second regime puts less energy in the weakest
combination of the first, however chaotic it is. This is why $\lambda_{\min}(M)$ need
not follow chaos intensity (Appendix~\ref{si:normalized-moment}).

\subsection{Fixed-point identifiability: a sparse-recovery argument}
\label{si:spark}

At a fixed point, $M$ has rank one and $\lambda_{\min}(M)=0$. A zero eigenvalue alone
does not rule out sparse recovery: sparse-recovery guarantees such as compressed sensing
and the $\ell_1$ recovery of \citet{tran2017exact} do not need a full-rank $M$, only one
that is non-degenerate on sparse differences. We show that fixed-point data fail even
this weaker condition: every sparse coefficient vector has a distinct, equally sparse or
sparser competitor that gives the same data. This holds for any dictionary, including
the oracle library, which explains why even the oracle prior does not rescue R0.

\medskip
\refstepcounter{proposition}\label{prop:spark}\noindent\textbf{Proposition~\theproposition} (Fixed-point spark collapse). Let $x^*$ be an
fixed point and suppose the observed states satisfy $x_n=x^*$ exactly for every
$n=0,\ldots,N-1$; equivalently, sample from the invariant measure
$\mu=\delta_{x^*}$. Let $\Phi=(\phi_1,\ldots,\phi_p)^\top$ be any dictionary
of $p\geq2$ functions with at least two terms nonzero at $x^*$ (true for every dictionary
considered in this paper: the constant term alone guarantees one, and any polynomial
dictionary of degree $\geq1$ in a nonzero variable, or with a nonzero $x^*$, guarantees a
second). Then the design matrix $\Theta\in\mathbb{R}^{N\times p}$ has
$\operatorname{spark}(\Theta)\leq2$. Moreover, for every nonzero coefficient vector
$\xi$ with sparsity $s=\|\xi\|_0$, there is a distinct $\xi'$ with
$\|\xi'\|_0\leq s$ and $\Theta\xi'=\Theta\xi$. Hence $\xi$ is not uniquely
recoverable from these data over the admissible sparsity class
$\mathcal C_s=\{z:\|z\|_0\leq s\}$.

\smallskip
\noindent\textit{Proof.} Every row of $\Theta$ equals $\Phi(x^*)^\top$, since $x_t=x^*$ for
every $t$ in the window; consequently $\mathrm{rank}(\Theta)=1$ (every row is a scalar
multiple of every other), which is the rank-1 structure shown for the fixed-point moment
matrix in Fig.~\ref{fig:fMM}A. Because $\Phi(x^*)\neq0$ (by hypothesis at least two dictionary
functions are nonzero there), a vector $w\in\mathbb{R}^p$ satisfies $\Theta w=\mathbf1_N
(\Phi(x^*)^\top w)=0$ exactly when the single scalar $\Phi(x^*)^\top w$ vanishes; the null
space of $\Theta$ is therefore the entire hyperplane
\[
  \ker(\Theta)\;=\;\{w\in\mathbb{R}^p:\Phi(x^*)^\top w=0\},
\]
of dimension $p-1$, a fact used below. Write
$V=\{i:\phi_i(x^*)\neq0\}$; by hypothesis $|V|\geq2$.

Picking any $j,k\in V$ and setting $v_j=\phi_k(x^*)$, $v_k=-\phi_j(x^*)$, $v_i=0$ otherwise
gives one explicit nonzero, $2$-sparse element of $\ker(\Theta)$ (direct check: $\Phi(x^*)^\top
v=\phi_j(x^*)\phi_k(x^*)-\phi_k(x^*)\phi_j(x^*)=0$), exhibiting two linearly dependent columns
of $\Theta$ directly, so $\operatorname{spark}(\Theta)\leq2$ (the spark is the smallest number
of linearly dependent columns). A small spark alone shows only that \emph{some} pair of
sparse vectors gives identical data~\citep{donoho2003optimally}. We now show that at a
fixed point \emph{every} sparse vector has such a competitor, which is possible because
$\ker(\Theta)$ is a whole hyperplane.

Fix any $\xi\in\mathbb{R}^p$ with $s:=\|\xi\|_0\geq1$ and let
$I=\operatorname{supp}(\xi)$. Because $|V|\geq2$, exactly one of three cases holds.

\emph{Case (I): $I\cap V=\varnothing$.} Every active term of $\xi$ vanishes at $x^*$, so
$\Theta\xi=\mathbf1_N\bigl(\Phi(x^*)^\top\xi\bigr)=\mathbf1_N\sum_{i\in I}\phi_i(x^*)\xi_i=0$.
The zero vector $\xi'=0$ then satisfies $\Theta\xi'=0=\Theta\xi$ with $\|\xi'\|_0=0<s$:
$\xi$ is not even the sparsest explanation of its own data, let alone the unique one.

\emph{Case (II): $I\cap V\neq\varnothing$ and $V\setminus I\neq\varnothing$.} Pick $i_0\in
I\cap V$ and $j_0\in V\setminus I$. Define $\xi'$ by leaving every other coordinate
unchanged and setting
\[
\xi'_{i_0}=0,
\qquad
\xi'_{j_0}=\frac{\phi_{i_0}(x^*)}{\phi_{j_0}(x^*)}\,\xi_{i_0}.
\]
Then $\Phi(x^*)^\top\xi'=\Phi(x^*)^\top\xi$, so
$\Theta\xi'=\Theta\xi$. Moreover,
$\operatorname{supp}(\xi')=(I\setminus\{i_0\})\cup\{j_0\}$: exactly $s$-sparse, and
$\xi'\neq\xi$ (they differ at $i_0$). This is a distinct, equally sparse alternative
that generates identical data.

\emph{Case (III): $V\subseteq I$.} Since $|V|\geq2$, choose a fixed nonzero
$v\in\ker(\Theta)$ supported on two indices in $V\subseteq I$. For any
$\tau\neq0$, set $\xi'=\xi+\tau v$. Then
$\operatorname{supp}(\xi')\subseteq I$, so $\|\xi'\|_0\leq s$,
$\Theta\xi'=\Theta\xi$, and $\xi'\neq\xi$.

In every case we have exhibited an explicit $\xi'\neq\xi$ with $\|\xi'\|_0\leq s$ and
$\Theta\xi'=\Theta\xi$: $\xi$ is never the unique explanation, at its own sparsity or
better, of the data it generates. Since $\xi$ was an arbitrary vector with $\|\xi\|_0\geq1$,
no coefficient vector with even a single active term can be uniquely identified from
exact fixed-point data by a data-only algorithm that treats all equally sparse
dictionary models as admissible. An external physical prior may restore uniqueness by
excluding every constructed competitor.
\hfill$\square$

\smallskip
\noindent\textbf{Scope.} Proposition~\ref{prop:spark} concerns the rank-one geometry of
a fixed point and the class of equally sparse dictionary models; an external physical
constraint that excludes the constructed competitors can restore uniqueness.
Proposition~\ref{prop:equivalence} covers exact collinearity of any subset of terms in any
regime.

\subsection{Finite-time concentration of the moment matrix}
\label{si:finite-time}

The convergence $\widehat M_N\to M$ of Appendix~\ref{app:sindy} gives no rate. The
following bound gives one when the dictionary products decorrelate along the trajectory,
and shows which sample size enters.

\medskip
\refstepcounter{proposition}\label{prop:finite-time}\noindent\textbf{Proposition~\theproposition} (Finite-time moment-matrix error).
Let $x_n=x(n\Delta t)$ with $x_0\sim\mu$, so the sampled process is stationary, and let
$\widehat M_N=\frac1N\sum_{n=0}^{N-1}\Phi(x_n)\Phi(x_n)^\top$ and
$M=\int\Phi\Phi^\top d\mu$. For every pair $(i,j)$, let
\[
f_{ij}(x)=\phi_i(x)\phi_j(x)-M_{ij},
\qquad
C_{ij}(k)=\mathbb E_\mu[f_{ij}(x_0)f_{ij}(x_k)],
\]
and assume that every $f_{ij}$ is square-integrable and
\[
A_{ij}=|C_{ij}(0)|+2\sum_{k=1}^\infty|C_{ij}(k)|<\infty.
\]
Then for every $\delta\in(0,1)$, with probability at least $1-\delta$ over
$x_0\sim\mu$,
\[
\|\widehat M_N-M\|_2
\leq
\sqrt{\frac{\sum_{i,j}A_{ij}}{\delta N}},
\]
and by Weyl's inequality the same bound holds for
$|\lambda_{\min}(\widehat M_N)-\lambda_{\min}(M)|$.

\smallskip
\noindent\textit{Proof.} Put $Y_{ij,N}=(\widehat M_N-M)_{ij}$. Stationarity and counting
pairs by their lag give
\[
\mathbb E_\mu|Y_{ij,N}|^2
=\frac1{N^2}\sum_{m,n=0}^{N-1}C_{ij}(|m-n|)
\leq\frac{A_{ij}}{N}.
\]
Summing over the finite set of entries gives
$\mathbb E_\mu\|\widehat M_N-M\|_F^2\leq\frac1N\sum_{i,j}A_{ij}$. Markov's inequality
for the squared Frobenius norm and $\|D\|_2\leq\|D\|_F$ give the bound.
\hfill$\square$

\smallskip
\noindent\textbf{Effective sample size.} Writing $A_{ij}=C_{ij}(0)\,\tau_{ij}$ when
$C_{ij}(0)>0$ shows that $N/\tau_{ij}$ acts as the effective sample size of each moment
entry: samples closer than the correlation time add little information. This is the
$N_{\mathrm{eff}}=N/\tau_{\mathrm{int}}$ used in the data budget $\mathcal B$
(Appendices~\ref{app:information-support} and~\ref{si:stats-hygiene}). Sampling a fixed
time interval more densely therefore does not create unlimited information. Summable
autocovariances are an assumption, not a consequence of chaos or of a positive Lyapunov
exponent.

\subsection{Transient data and identifiability}
\label{si:transient}

The effect of transient or perturbation data depends on whether those observations
are appended or replace settled observations. The following statement keeps total
information, per-sample information, and the asymptotic information rate separate.

\medskip
\refstepcounter{proposition}\label{prop:transient}\noindent\textbf{Proposition~\theproposition} (Transient observations and weakest-direction
information). Let $\Theta_{\mathrm{attr}}\in\mathbb R^{N_{\mathrm{attr}}\times p}$
be an attractor design and $\Theta_{\mathrm{trans}}\in\mathbb
R^{N_{\mathrm{trans}}\times p}$ a finite block of transient or perturbation rows.
Define
\[
G_{\mathrm{attr}}=\Theta_{\mathrm{attr}}^\top\Theta_{\mathrm{attr}},
\quad
G_{\mathrm{trans}}=\Theta_{\mathrm{trans}}^\top\Theta_{\mathrm{trans}},
\quad
G_{\mathrm{all}}=G_{\mathrm{attr}}+G_{\mathrm{trans}}.
\]
Then:
\begin{enumerate}
  \item[(i)] Appending the transient block cannot decrease total weakest-direction
  information:
  \[
  \lambda_{\min}(G_{\mathrm{all}})\geq
  \lambda_{\min}(G_{\mathrm{attr}}),
  \qquad
  \sigma_{\min}\!\begin{pmatrix}\Theta_{\mathrm{attr}}\\
  \Theta_{\mathrm{trans}}\end{pmatrix}
  \geq\sigma_{\min}(\Theta_{\mathrm{attr}}).
  \]
  The inequality is strict exactly when
  \[
  E_{\min}(G_{\mathrm{attr}})\cap\ker(G_{\mathrm{trans}})=\{0\},
  \]
  where $E_{\min}$ is the minimum-eigenvalue eigenspace. If the attractor design is
  singular, the stacked design has full column rank exactly when
  $\ker(\Theta_{\mathrm{attr}})\cap
  \ker(\Theta_{\mathrm{trans}})=\{0\}$.

  \item[(ii)] Under a fixed total budget, replacing a fraction $\tau$ of attractor
  rows by transient rows gives the normalized mixture
  \[
  \widehat M(\tau)=(1-\tau)\widehat M_{\mathrm{attr}}
  +\tau\widehat M_{\mathrm{trans}},\qquad0\leq\tau\leq1,
  \]
  and
  \[
  \lambda_{\min}(\widehat M(\tau))\geq
  (1-\tau)\lambda_{\min}(\widehat M_{\mathrm{attr}})
  +\tau\lambda_{\min}(\widehat M_{\mathrm{trans}}).
  \]
  Hence $\lambda_{\min}(\widehat M_{\mathrm{trans}})\geq
  \lambda_{\min}(\widehat M_{\mathrm{attr}})$ is sufficient for non-decrease, but
  not necessary. If the smallest eigenvalue of $\widehat M_{\mathrm{attr}}$ is simple
  with unit eigenvector $u$, then
  \[
  \left.\frac{d}{d\tau}\lambda_{\min}(\widehat M(\tau))\right|_{\tau=0}
  =u^\top\widehat M_{\mathrm{trans}}u
  -\lambda_{\min}(\widehat M_{\mathrm{attr}}).
  \]
  This derivative is only a local criterion; the weakest eigendirection may rotate as
  $\tau$ grows.

  \item[(iii)] Let $N_{\mathrm{trans}}$ remain fixed while the attractor tail grows,
  and assume that this tail satisfies the typicality and integrability hypotheses of
  the Appendix~\ref{app:sindy} moment-convergence result for an ergodic measure $\mu$. With
  \[
  M_{\mathrm{attr}}=\int\Phi(x)\Phi(x)^\top\,d\mu(x),
  \]
  the full normalized empirical moment obeys
  $\widehat M_N\to M_{\mathrm{attr}}$ and therefore
  $\lambda_{\min}(\widehat M_N)\to\lambda_{\min}(M_{\mathrm{attr}})$.
\end{enumerate}

\smallskip
\noindent\textit{Proof.}
For (i), $v^\top G_{\mathrm{trans}}v=\|\Theta_{\mathrm{trans}}v\|_2^2\geq0$, so
$G_{\mathrm{all}}\succeq G_{\mathrm{attr}}$ and Courant--Fischer gives
non-decrease. Equality holds if a nonzero minimum direction of
$G_{\mathrm{attr}}$ lies in $\ker(G_{\mathrm{trans}})$. Conversely, if the minimum
eigenvalues are equal, a unit minimizer of $G_{\mathrm{all}}$ must attain the minimum
of $G_{\mathrm{attr}}$ and contribute zero to the nonnegative transient quadratic
form, proving the intersection criterion. The singular full-rank statement follows
from the kernel of the vertically stacked design.

For (ii), apply the Rayleigh characterization to each unit vector and then minimize:
$v^\top\widehat M(\tau)v\geq(1-\tau)\lambda_{\min}
(\widehat M_{\mathrm{attr}})+\tau\lambda_{\min}
(\widehat M_{\mathrm{trans}})$. The derivative is the standard
Hellmann--Feynman formula for a simple eigenvalue.

For (iii), write
\[
\widehat M_N=\frac{N_{\mathrm{attr}}}{N}\widehat M_{\mathrm{attr},N}
+\frac{N_{\mathrm{trans}}}{N}\widehat M_{\mathrm{trans}}.
\]
The Appendix~\ref{app:sindy} result gives $\widehat M_{\mathrm{attr},N}\to M_{\mathrm{attr}}$,
while $N_{\mathrm{attr}}/N\to1$ and
$(N_{\mathrm{trans}}/N)\widehat M_{\mathrm{trans}}\to0$. Matrix and eigenvalue
continuity give the claim.
\hfill$\square$

\begin{figure}[t!]
\centering
\includegraphics[width=\linewidth]{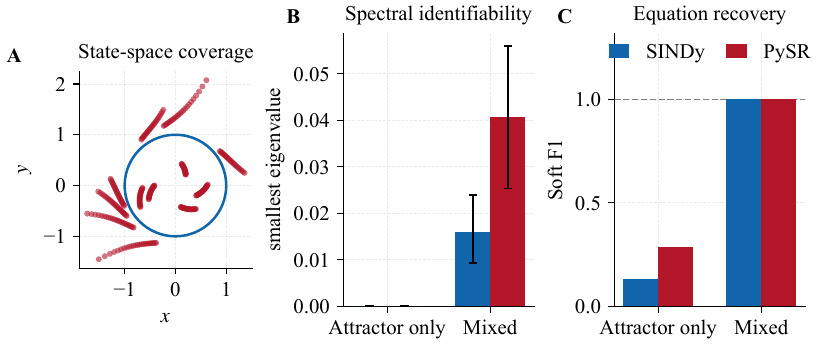}
\caption{\textbf{Matched-budget transient observations resolve an exact attractor
degeneracy.} For the stable Hopf normal form, the degree-three polynomial library has
three pointwise identities on the unit-circle attractor and hence rank seven rather than
ten. Each design contains 600 exact state/analytic-derivative pairs. The attractor-only
design remains singular and admits many observationally equivalent vector fields. Replacing
half the samples by early transient observations from varied initial radii restores full
rank, gives positive full-library and residualized spectral scales, and yields exact SINDy
and PySR recovery in the recorded noiseless runs. Values summarize three independently
seeded designs; attractor-only algorithm scores are reference selections among nonunique
solutions, not information-theoretic floors. (\textit{A})~Unit-circle attractor (blue) and
transient samples (red). (\textit{B})~Smallest eigenvalue of the full-library moment matrix
$\widehat M$ (SINDy, blue) and $\sigma_{\mathrm{sep}}^2$, the smallest eigenvalue of the
residualized true-term moment matrix used by PySR (red; Appendix~\ref{app:pysr}); bars give
the mean and error bars the range over the three designs. (\textit{C})~Soft~F1.}
\label{fig:si-transient-demo}
\end{figure}

\medskip
\noindent\textbf{Interpretation.}
Data off the attractor can excite dictionary combinations that vanish on it, but being
off the attractor does not guarantee this. By (i), appended transient data raise the
weakest-direction information exactly when they have energy along every weakest
direction of the attractor design. By (iii), a fixed transient block can make every
finite design full rank, but its share of $\widehat M_N$ decays as $1/N$; only a
positive fraction of informative off-attractor data keeps the information per sample
positive. Figure~\ref{fig:si-transient-demo} shows the mechanism on the Hopf normal
form. Our main experiments discard burn-in, so they do not test the use of transients.

\section{Jerk systems: a non-polynomial held-out family}\label{app:jerk}
\label{si:jerk}

L84 and L96 share a structural feature that could be read as favouring the framework:
both are degree-2/3 polynomial systems, and both SINDy libraries already contain the
ground-truth terms. This section applies $\lambda_{\min}(M)$, $\mathcal I_{\mathrm{SINDy}}$, and
$\mathcal{F}_{\mathrm{PySR}}$, unmodified, to a third, unrelated ODE family with two
genuinely non-polynomial members, to test whether the regime-ordering story is an
artifact of the polynomial setting or transfers to a harder functional form.

\emph{Systems.} All three jerk systems share the form
$\dot{x}=y,\ \dot{y}=z,\ \dot{z}=-0.9z-y-bx+F(x,y)$, with $b$ the control parameter and
$F$ varying by system: \textbf{poly}, $F=xy^2$ (the in-dictionary control, identical in
character to L84/L96); \textbf{rat}, $F=xy^2/(1+(xy)^2/64)$; \textbf{sinrat},
$F=(xy^2+\sin x)/(1+(xy)^2/64)$. rat and sinrat are genuinely non-polynomial: no
finite-degree polynomial dictionary contains their true right-hand side.
Each system has the same 8-regime structure as L84/L96 (R0 fixed point; R1--R2 limit
cycle; R3--R7 chaos, $\lambda_1$-binned), with 5 trajectory slots per regime.

\emph{Sample sizes.} All planned fits completed successfully.
For each system, PySR was fitted across 8 regimes, 5 slots, and 5 random
initialisations, yielding 200 fits for the scored $\dot{z}$ equation
and 600 fits across all three state equations (poly, rat, and sinrat).
Regime--slot values are reported as averages over the 5 initialisations
(40 regime--slot cells per system).
SINDy fits are likewise complete, with 40 fits per system and equation
(8 regimes $\times$ 5 slots).

\emph{Library and operators.} SINDy is given the degree-3 polynomial basis (as in L84/L96) plus the
system's own ground-truth nonlinear term as one additional feature column (e.g., for rat, the
column $xy^2/(1+(xy)^2/64)$ itself), the same ``ground-truth term included'' convention as
the main experiments' default library, not the separate null/oracle prior-quality
manipulation. PySR is given the corresponding term as a custom operator. Only the
$\dot{z}$ equation is scored: $\dot{x}=y$ and $\dot{y}=z$ are exact linear identities in
every regime and carry no information about discoverability.

\emph{Protocol.} Each regime and slot uses one noiseless training window of $N=9{,}000$
samples with exact derivatives; there is no data-volume or noise sweep for these
systems. Consequently $Q_{\mathrm{noise}}=1$ and
$\mathcal{F}_{\mathrm{PySR}}=\sqrt{\sigma_{\mathrm{sep}}^{1/4}}$. All quantities are
computed from the raw trajectories.

\begin{figure}[ht]
\centering
\includegraphics[width=\linewidth]{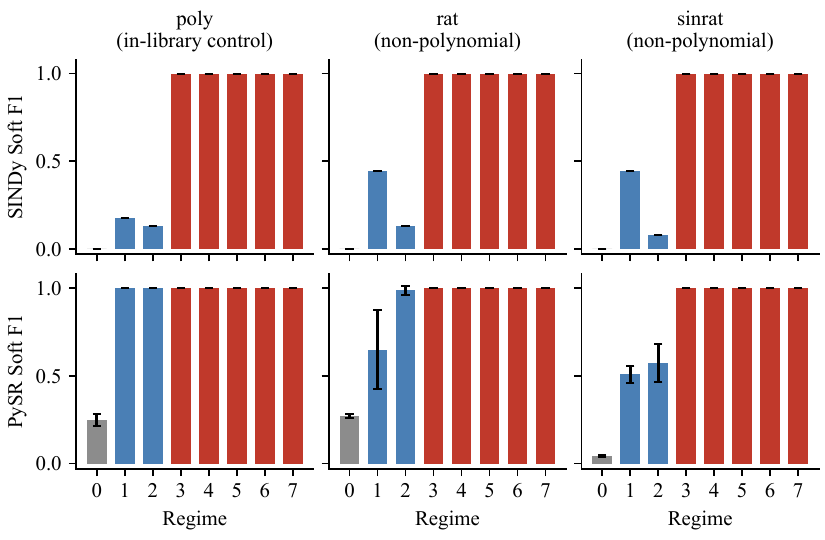}
\caption{\textbf{Regime ordering on the jerk systems.}
Mean Soft~F1 across 5 trajectory slots per regime
(error bars: SD across slots), for SINDy (top) and PySR (bottom),
on poly (in-dictionary control) and the two non-polynomial systems
rat and sinrat.
Grey~=~fixed point, blue~=~limit cycle (R1--R2), red~=~chaos (R3--R7).
PySR values average across the 5 initialisations of each slot
(200 dim-2 fits per system).}
\label{fig:si-jerk}
\end{figure}

\emph{Results.} Figure~\ref{fig:si-jerk} shows the same qualitative pattern as L84 and
L96 in all three systems: the fixed point is lowest and the limit cycles lie below the
chaotic regimes. SINDy scores Soft~F1~$=0$ at R0, $0.08$--$0.44$ at R1 and R2, and
$\approx1.0$ from R3 on, including in the two non-polynomial systems. PySR shows the same
floor-then-ceiling pattern for poly, and a graded ordering R0~$<$~R1~$<$~R2~$<$~chaos for
rat ($0.27$, $0.65$, $0.99$, $1.0$) and sinrat ($0.05$, $0.51$, $0.57$, $1.0$). Spearman
$\rho$ between each indicator and Soft~F1, pooled over all 8 regimes $\times$ 5 slots
($n = 40$ per system for both SINDy and PySR):

\begin{center}
\begin{tabular}{lccc}
\hline
 & poly & rat & sinrat \\
\hline
$\mathcal I_{\mathrm{SINDy}}$ vs.\ Soft~F1 & $0.87$ & $0.86$ & $0.84$ \\
$\mathcal{F}_{\mathrm{PySR}}$ vs.\ Soft~F1  & $0.57$ & $0.58$ & $0.75$ \\
\hline
\end{tabular}
\end{center}

Pooled across all three systems ($n = 120$ per method), $\rho = 0.85$ for
$\mathcal{I}_{\mathrm{SINDy}}$ and $\rho = 0.61$ for $\mathcal{F}_{\mathrm{PySR}}$.

\emph{Why the PySR correlation is lower here.} Table~\ref{tab:noiseless} pools a
data-volume sweep, a noise sweep, and a prior-quality sweep, and each adds variation in
the outcome. The jerk check is a single noiseless condition. Under the same restriction
(one noiseless $N$, taken from the data-volume experiment), the PySR correlation on L84
and L96 is $0.606$ and $0.609$, close to poly ($0.57$) and rat ($0.58$) and below sinrat
($0.75$). In this design, many non-fixed-point cells already reach Soft~F1~$=1$ under
PySR (all of them for poly, $86\%$ for rat, $71\%$ for sinrat), so Spearman's $\rho$ has
little outcome variation left to explain. $\mathcal I_{\mathrm{SINDy}}$ stays at
$0.84$--$0.87$ under the same restriction, because SINDy's Soft~F1 is low at the fixed
point and the limit cycles and high in chaos, in the same order as its indicator. The lower PySR value therefore reflects the single-condition design, not a
difference between polynomial and non-polynomial systems. For rat, part of the gap comes
from the PySR search itself: at R1 the 25 fits split between the exact rational term
(9 fits, Soft~F1~$\approx1.0$), a different rational form built from $x_1^3$ and $x_1^2x_2$
(12 fits, $0.26$--$0.37$), and intermediate forms (4 fits, $0.69$--$0.76$). Averaging the
slots of each regime before correlating removes this search variability and gives
$\rho=0.79$ for rat.

\emph{Scope.} This is a cross-regime validation at a single noiseless $N$, not the full
data-volume, noise, and prior-quality replication reported for L84 and L96 in
Table~\ref{tab:noiseless}.

\section{Additional results and statistics}\label{app:l96}

\subsection{Moment-matrix spectrum across regimes, raw and normalized}
\label{si:normalized-moment}

Because $\lambda_{\min}(M)$ is a Gram-matrix eigenvalue, it is not scale-invariant: an
attractor with the same correlation structure but larger amplitude has a
proportionally larger $\lambda_{\min}(M)$. To check that the regime trends of
$\lambda_{\min}(M)$ reflect a change in dictionary excitation rather than attractor
amplitude alone, we recomputed $\lambda_{\min}$ on the
correlation-normalized moment matrix $D^{-1/2}MD^{-1/2}$ ($D=\mathrm{diag}(M)$, unit
variance on every dictionary column), using the same degree-3 polynomial dictionary
($p{=}20$ for L84, $p{=}56$ for L96) and the same per-regime average over 5
trajectory slots as the main-text $\lambda_{\min}(M)$.

\begin{center}
\begin{tabular}{llrrr}
\hline
System & Regime & $\lambda_1$ (ref.) & $\lambda_{\min}(M)$ & $\lambda_{\min}(D^{-1/2}MD^{-1/2})$ \\
\hline
L84 & R0 (fp)  & $-0.150$ & $\approx 0$            & $\approx 0$ \\
L84 & R1 (lc)  & $0.000$  & $6.31\times10^{-5}$     & $9.15\times10^{-5}$ \\
L84 & R2 (lc)  & $0.000$  & $1.60\times10^{-3}$     & $8.10\times10^{-4}$ \\
L84 & R3 (chaos) & $0.073$ & $3.29\times10^{-3}$    & $1.10\times10^{-3}$ \\
L84 & R4 (chaos) & $0.142$ & $4.98\times10^{-3}$    & $1.63\times10^{-3}$ \\
L84 & R5 (chaos) & $0.168$ & $1.96\times10^{-3}$    & $8.05\times10^{-4}$ \\
L84 & R6 (chaos) & $0.189$ & $4.75\times10^{-3}$    & $1.49\times10^{-3}$ \\
L84 & R7 (chaos) & $0.207$ & $1.70\times10^{-3}$    & $7.57\times10^{-4}$ \\
\hline
L96 & R0 (fp)  & $-0.300$ & $\approx 0$            & $\approx 0$ \\
L96 & R1 (lc)  & $0.000$  & $\approx 0$             & $\approx 0$ \\
L96 & R2 (lc)  & $0.000$  & $\approx 0$             & $\approx 0$ \\
L96 & R3 (chaos) & $0.507$ & $1.38\times10^{-3}$    & $2.97\times10^{-4}$ \\
L96 & R4 (chaos) & $0.754$ & $3.80\times10^{-3}$    & $7.46\times10^{-4}$ \\
L96 & R5 (chaos) & $0.932$ & $6.89\times10^{-3}$    & $1.21\times10^{-3}$ \\
L96 & R6 (chaos) & $1.133$ & $9.87\times10^{-3}$    & $1.72\times10^{-3}$ \\
L96 & R7 (chaos) & $1.348$ & $1.26\times10^{-2}$    & $2.42\times10^{-3}$ \\
\hline
\end{tabular}
\end{center}

The fixed-point and limit-cycle values reported as $\approx 0$ are at the numerical
eigenvalue floor ($10^{-14}$--$10^{-16}$, occasionally slightly negative from
floating-point error on a rank-deficient positive-semidefinite matrix), consistent
with $\lambda_{\min}=0$ exactly rather than a real ordering violation.

Two results survive normalization unchanged. First, the hardest regimes stay separated.
In both systems and under both matrices, the fixed point sits at the numerical floor and
every chaotic regime lies well above it. In L84, the low-amplitude limit cycle R1 lies
roughly one to two orders of magnitude below the other regimes, while the large-amplitude
limit cycle R2 falls at or within the chaotic range. In L96, both limit cycles sit at the
numerical floor for this dictionary.
Second, the \emph{within-chaos rank ordering} is identical between the raw and
normalized columns for every regime in both systems. Using two-sided exact
permutation tests for the five chaotic regimes gives
$\rho(\lambda_1,\lambda_{\min})=-0.50$ for L84
($n=5$, $p=0.4500$) and $\rho=+1.00$ for L96
($n=5$, $p=0.0167$), under both normalizations.

Because correlation normalization removes the scale of each feature while preserving
these orderings, the trends are not an effect of feature amplitude alone. They are
empirical: neither the regime label nor $\lambda_1$ orders moment matrices
(Remark~3 in Appendix~\ref{si:exact-collinearity}). SINDy is fit on the raw library, so
the raw $\lambda_{\min}(M)$ describes its regression design; the PySR indicator normalizes
the candidate terms before residualization (Appendix~\ref{app:pysr}).

\subsection{PySR partial correlations beyond the noise-level ordering}
\label{si:partial-correlation}

In the noise experiment, both $\mathcal F_{\mathrm{PySR}}$ and Soft~F1 decrease as the noise
ratio $\eta$ grows. Their correlation could therefore reflect the noise level alone,
without the regime doing any work. We test this in two ways, with the convention of
Table~\ref{tab:noiseless}: Soft~F1 is averaged over the PySR initializations of each slot,
and correlations are computed for each equation and then averaged over equations. The
partial Spearman correlation removes the effect of $\eta$ and of the derivative track
from the ranks of both quantities. The within-stratum correlation fixes $\eta$ and the
track exactly (7 noise ratios $\times$ 2 tracks $=$ 14 strata), so that only the regime and
the slot vary (4 regimes $\times$ 5 slots $=$ 20 trajectories per equation and stratum).
Confidence intervals come from 1000 trajectory resamples.

\begin{center}
\small
\begin{tabular}{lcccc}
\hline
System & $\rho$ (Table~\ref{tab:noiseless}) & Partial $\rho$ ($\eta$, track) & Within stratum, $\eta>0$ & Within stratum, $\eta=0$ \\
 & & [95\% CI] & median (min--max) & FD, SG \\
\hline
L84 & $0.81$ & $0.58$ $[0.26,\,0.73]$ & $0.45$ ($0.13$--$0.59$) & $-0.22$, $-0.32$ \\
L96 & $0.88$ & $0.82$ $[0.69,\,0.88]$ & $0.72$ ($0.49$--$0.90$) & $0.44$, $0.43$ \\
\hline
\end{tabular}
\end{center}

On L96, the correlation stays strong under both controls. On L84, it weakens but stays
positive: the partial correlation is $0.58$, and the within-stratum correlation is
positive in all twelve strata with noise, between $0.28$ and $0.59$ for $\eta\geq0.05$.
The indicator therefore ranks the regimes at a fixed noise level, not only the noise
levels. The exception is zero noise, where $Q_{\mathrm{noise}}=1$ and only
$\sigma_{\mathrm{sep}}$ varies. The four L84 regimes then all reach Soft~F1 between $0.85$
and $0.96$, and $\sigma_{\mathrm{sep}}$ does not rank them.

The per-equation convention matters on L84. Pooling the three equations into one
correlation per stratum gives a median near zero, because the equations differ in level
in opposite directions. Under noise ($\eta\geq0.05$), the $\dot x_0$ equation has the lowest
mean indicator ($0.04$, against $0.10$ and $0.11$) but the highest mean Soft~F1 ($0.65$,
against $0.33$ and $0.39$). $\mathcal F_{\mathrm{PySR}}$ therefore ranks experimental
conditions within an equation, but its level is not comparable across equations.

\subsection{L96 prior-quality library construction (SINDy)}
The L96 prior-quality experiment uses the same null/overcomplete/oracle scheme as
L84 (Appendix~\ref{app:experimental-details}), constructed against the L96 right-hand side
$\dot{x}_i = (x_{i+1}-x_{i-2})x_{i-1} - x_i + F$ (5-dimensional, cyclic indices).
The null library contains 10 cubic decoys with no linear or bilinear structure:
$x_i^3$ for $i=0,\ldots,4$ and $x_i^2 x_{i+1}$ (cyclic) for $i=0,\ldots,4$, none of
which coincide with a ground-truth term.
The overcomplete library is the full degree-3 polynomial library with bias
(56 features for 5 variables, containing all 16 ground-truth terms among 40 distractors).
The oracle library is the 16 exact ground-truth terms: the constant $1$, the 5 linear
terms $x_i$, and the 10 bilinear cross-pairs $x_i x_j$ ($i<j$) spanning the terms
that appear in the expanded right-hand side, and nothing else.
As in L84, the oracle condition is an intentional upper-anchor isolating
coefficient estimation from term selection, not a realistic operating point.

\subsection{Soft~F1: implementation conventions and worked examples}
\label{si:softf1-definition}

This subsection fixes the conventions behind Eq.~\ref{eq:softf1}. Both the ground-truth and the discovered
expression are prepared identically: (i) the expression is expanded; (ii) floating-point constants within
$10^{-3}$ of a simple rational are snapped to it, so that, for instance, $\log(x_0^2+1.0000081)$ and
$\log(x_0^2+1)$ define the same structural term; (iii) summands of magnitude below $10^{-6}$ are removed as
numerical noise; and (iv) additive terms that share a non-trivial denominator are recombined, and each term is
factored so that a scalar prefactor does not change its structural key. Terms with the same structural key are
summed. The term sets $\mathcal{T}$ and $\mathcal{D}$ contain the terms whose coefficient magnitude exceeds
$10^{-8}$. No projection onto a fixed dictionary is used: a discovered term outside the polynomial
dictionary of the ground truth (for example an extra $\sin x_0$ returned by symbolic regression) belongs to
$\mathcal{D}$ and lowers soft precision, exactly as a spurious monomial does.

Three conventions complete the definition. A discovered expression with no terms while the ground truth is
non-empty scores $0$. An expression that is algebraically identical to the ground truth, however written
(for instance $(A+B)/D$ versus $A/D+B/D$), scores exactly $1$. A missing or wholly spurious term has
$\delta_j=1$ and hence $s_j=e^{-3}\approx0.050$ inside the relevant average; a coefficient with the correct
magnitude but the wrong sign has $\delta_j=1$ as well.

Worked examples for the ground truth $x_0x_1-4x_0x_2-x_1+1$ (four terms):
\begin{center}
\begin{tabular}{lccc}
\hline
Discovered expression & $P_{\mathrm{soft}}$ & $R_{\mathrm{soft}}$ & Soft~F1 \\
\hline
exact recovery & $1$ & $1$ & $1.000$ \\
$-4x_0x_2\to-2x_0x_2$ ($s_j=e^{-1}$) & $0.842$ & $0.842$ & $0.842$ \\
constant term missing & $1$ & $0.762$ & $0.865$ \\
spurious term $+x_2$ & $0.810$ & $1$ & $0.895$ \\
constant missing and $+x_2$ spurious & $0.762$ & $0.762$ & $0.762$ \\
empty expression ($\mathcal{D}=\emptyset$) & undefined & $0.050$ & $0$ (by convention) \\
\hline
\end{tabular}
\end{center}

\subsection{Statistics: Soft~F1 \texorpdfstring{$\alpha$}{alpha} sensitivity, effective sample size, and
bootstrap intervals}
\label{si:stats-hygiene}

\emph{Soft~F1 $\alpha$ sensitivity.} The coefficient fidelity in Soft~F1,
$s_j=\exp(-\alpha\delta_j)$, uses $\alpha=3$ throughout. Recomputing Soft~F1 directly from the
stored discovered-expression strings at $\alpha\in\{1.5,3,6\}$ (data-volume experiment, both
algorithms, both systems, at a representative small and the largest $N$) leaves the
regime ordering unchanged: the Spearman correlation between regime-mean Soft~F1 at
$\alpha=3$ and at $\alpha=1.5$ or $\alpha=6$ is $1.000$ in 14 of 16 (system, algorithm,
$N$, $\alpha$) comparisons, and $0.976$ in the other two (L84 SINDy, small $N$, $\alpha=3$ vs.\ $\alpha=6$;
L96 SINDy, small $N$, $\alpha=3$ vs.\ $\alpha=1.5$), each a single near-tied pair of regimes swapping rank, not a qualitative change.

\emph{Effective sample size.} Trajectory samples within a regime are autocorrelated;
we estimate the integrated autocorrelation time $\tau_{\mathrm{int}}$ per regime (averaged
over dimensions and slots, from the clean reference trajectory) and report
$N_{\mathrm{eff}}=N/\tau_{\mathrm{int}}$. Limit-cycle regimes have $1.97\times$ (L84) to
$2.93\times$ (L96) longer $\tau_{\mathrm{int}}$ than chaotic regimes, so a given raw $N$
corresponds to substantially fewer independent samples in a limit cycle than in chaos.
Re-expressing the data-volume Soft~F1 curves against $N_{\mathrm{eff}}$ rather than raw $N$
does not change the ordering: in L96, chaotic regimes reach Soft~F1~$\approx1.0$ by
$N_{\mathrm{eff}}\approx30$--$45$, while the limit-cycle regime R2 at the \emph{same}
$N_{\mathrm{eff}}$ remains at Soft~F1~$\approx0.10$ (SINDy); in L84, R2 (the
high-amplitude limit cycle) reaches Soft~F1~$=1.0$ by $N_{\mathrm{eff}}\approx40$, matching
chaos, while R1 (the low-amplitude limit cycle) plateaus at Soft~F1~$\approx0.64$ even at
its maximum achievable $N_{\mathrm{eff}}\approx60$ (raw $N=10{,}000$), the plateau shown in
Fig.~\ref{fig:f2}. This is consistent with the R1/R2 distinction already drawn in the
main text: the limit-cycle disadvantage is not an autocorrelation/effective-sample-size
artifact, since at matched $N_{\mathrm{eff}}$, chaotic regimes succeed while the harder
limit-cycle regime does not.

\emph{Trajectory-level cluster bootstrap.} Cells that share a trajectory are not
independent. We therefore resample whole trajectories (regime$\times$slot) with
replacement and recompute the per-equation-averaged Spearman $\rho$ of
Table~\ref{tab:noiseless}, for both indicators.

\begin{table}[t!]
\centering
\caption{\textbf{Trajectory-level cluster bootstrap for both indicators.} Each row
resamples whole trajectories (regime$\times$slot) with replacement ($5000$ replicates)
and recomputes the per-equation-averaged $\rho$; $n_{\mathrm{traj}}$ is the number of
independent trajectories and the CI is the $95\%$ percentile interval. Every interval
excludes zero.}
\begin{tabular}{lllcc}
\hline
Score & System & Experiment & $n_{\mathrm{traj}}$ & $95\%$ CI \\
\hline
$\mathcal{I}_{\mathrm{SINDy}}$ & L84 & data volume   & $40$ & $[0.78,\,0.90]$ \\
$\mathcal{I}_{\mathrm{SINDy}}$ & L84 & noise         & $20$ & $[0.88,\,0.95]$ \\
$\mathcal{I}_{\mathrm{SINDy}}$ & L84 & prior quality & $40$ & $[0.90,\,0.94]$ \\
$\mathcal{I}_{\mathrm{SINDy}}$ & L96 & data volume   & $40$ & $[0.87,\,0.93]$ \\
$\mathcal{I}_{\mathrm{SINDy}}$ & L96 & noise         & $20$ & $[0.73,\,0.94]$ \\
$\mathcal{I}_{\mathrm{SINDy}}$ & L96 & prior quality & $40$ & $[0.88,\,0.93]$ \\
\hline
$\mathcal{F}_{\mathrm{PySR}}$  & L84 & data volume   & $40$ & $[0.69,\,0.85]$ \\
$\mathcal{F}_{\mathrm{PySR}}$  & L84 & noise         & $20$ & $[0.78,\,0.84]$ \\
$\mathcal{F}_{\mathrm{PySR}}$  & L84 & prior quality & $40$ & $[0.83,\,0.88]$ \\
$\mathcal{F}_{\mathrm{PySR}}$  & L96 & data volume   & $40$ & $[0.71,\,0.88]$ \\
$\mathcal{F}_{\mathrm{PySR}}$  & L96 & noise         & $20$ & $[0.83,\,0.90]$ \\
$\mathcal{F}_{\mathrm{PySR}}$  & L96 & prior quality & $40$ & $[0.87,\,0.94]$ \\
\hline
\end{tabular}
\label{tab:clusterboot}
\end{table}

\emph{Within-chaos bootstrap.} Each within-chaos correlation has only five
regime-level points. Bootstrapping over the five trajectory slots per regime
($5000$ resamples, recomputing each regime mean from a slot sample drawn with
replacement) probes slot-level variability conditional on these five regimes;
it does not quantify uncertainty over the selection of regimes. It gives a
wide $95\%$ interval for L84 ($[-0.90,-0.10]$), with every conditional
resample negative, and exactly $+1$ for L96. Thus the L84 point estimate is
stable to resampling the available slots, but the five-regime association is not
significant (exact permutation test, Appendix~\ref{si:normalized-moment}).

\subsection{Per-dimension Soft~F1}
\label{si:per-dimension}

Figure~\ref{fig:f2} and Table~\ref{tab:noiseless} average over the equations of each
system. For L96, whose five equations are cyclically identical, the per-equation mean
Soft~F1 differs by at most $0.021$ in the data-volume and noise experiments. For L84, whose
three equations differ, PySR recovers the $\dot x_0$ equation worst in the data-volume
experiment ($0.606$ against $0.829$ for $\dot x_1$ and $\dot x_2$) but best in the noise
experiment ($0.717$ against $0.566$).

\section{R1 hyperparameter control: noiseless recovery and noise collapse}\label{app:hyperparam-control}
\label{si:hyperparam-control}

The L84 R1 (limit-cycle, low $\lambda_{\min}(M)$) recovery ceiling reported in
Fig.~\ref{fig:f2} reflects the fixed STLSQ configuration rather than an
information-theoretic impossibility: R1 has $\lambda_{\min}(M)>0$
(no exact polynomial identity, unlike the L96 limit cycles), so noiseless recovery is possible in
principle. This section supplies the empirical demonstration.

\emph{Protocol.} SINDy (STLSQ, degree-3 polynomial library) was fit to L84 R1 (and R2, for
contrast) across a grid of ridge penalty $\alpha\in\{0.05,10^{-3},10^{-4},0\}$ crossed with
sparsity threshold $\lambda$: two fixed values ($0.05$, the deployment default, and $0.01$)
plus a diagnostic value $\lambda_{\mathrm{scaled}}=0.5\,c_{\min}\,w_{\min}(\sigma_{\min},\alpha)$,
where $c_{\min}=0.25$ is the smallest ground-truth L84 coefficient magnitude (the $-0.25\,x_0$ term)
and $w_{\min}=\sigma_{\min}^2/(\sigma_{\min}^2+\alpha)$ is the worst singular-direction
shrinkage. This diagnostic uses oracle coefficient knowledge and is not claimed to be an exact
feature-level survival threshold unless the relevant feature aligns with that singular direction.
Leg~1 (noiseless) used exact derivatives and swept $N$ over the same 17-point
log grid as the data-volume experiment ($20$ to $10{,}000$). Leg~2 (noisy) fixed
$N=5{,}000$, matching the noise experiment, and applied
additive Gaussian measurement noise at $\eta\in\{0.02,0.05\}$ with finite-difference
derivatives, following the same noise-injection protocol as the noise experiment
(Appendix~\ref{app:experimental-details}). Both legs average Soft~F1 over the 5 trajectory slots per regime.
\emph{Caveat:} this experiment uses a freshly regenerated draw of the R1/R2 trajectory
slots, independent of the draw underlying the data-volume curve of Fig.~\ref{fig:f2}; the
deployment-configuration baseline reported here (Soft~F1~$\approx0.64$ at $N=10{,}000$) is nevertheless
the same as the R1 plateau of Fig.~\ref{fig:f2} ($\approx0.64$), and the R1/R2 asymmetry central to the claim is
reproduced within this fresh draw as well. Scoring follows the rest of the paper: every fit is scored with
Eq.~\ref{eq:softf1} against the ground truth of its own trajectory, built from that trajectory's forcing $F$
(Appendix~\ref{app:experimental-details}), and the exact derivatives of the noiseless leg use the same $F$.

\begin{figure}[ht]
\centering
\includegraphics[width=\linewidth]{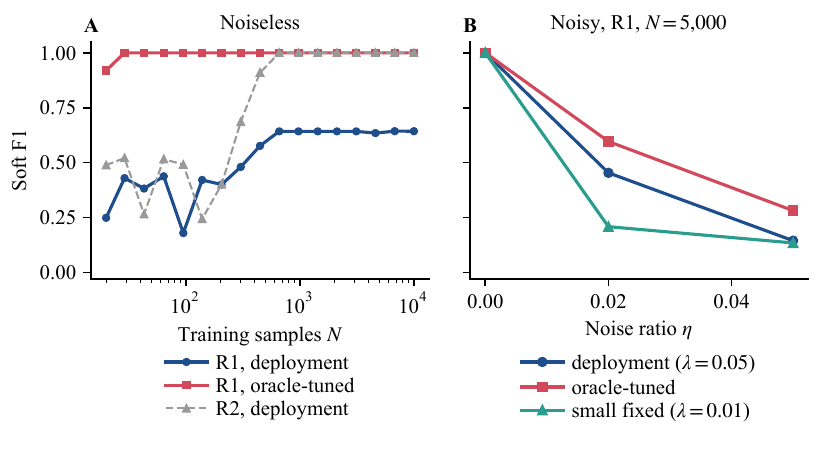}
\caption{\textbf{R1 hyperparameter control.} (A) Noiseless Soft~F1 vs.\ training-set size $N$
for L84 R1 at the deployment configuration ($\alpha=\lambda=0.05$, blue) vs.\ an oracle-tuned
configuration ($\alpha=0$, $\lambda$ scaled to $\sigma_{\min}$, red); R2 at the deployment
configuration shown for reference (grey, dashed). (B) Soft~F1 vs.\ measurement noise ratio
$\eta$ for R1. The noisy points use $N=5{,}000$; the $\eta=0$ reference is the nearest
noiseless grid point, $N=4{,}597$. The curves compare the deployment threshold, the oracle-tuned
threshold, and a small fixed threshold ($\lambda=0.01$), all at $\alpha=0$.}
\label{fig:si-hyperparam}
\end{figure}

\emph{Results, noiseless (Fig.~\ref{fig:si-hyperparam}A).} At the deployment configuration, R1
grows slowly and plateaus at Soft~F1~$\approx0.64$ by $N\approx700$, never reaching $1.0$ even
at $N=10{,}000$, the ceiling already reported in the main text. Setting $\alpha=0$ and
$\lambda=\lambda_{\mathrm{scaled}}$ reaches Soft~F1~$=1.0$ by $N\approx30$ and remains there for
every larger $N$: the ceiling is not intrinsic to R1's dynamics, it is a property of the fixed
threshold/ridge configuration relative to what $\lambda_{\min}(M)$ can resolve. R2 (higher
$\lambda_{\min}(M)$) reaches Soft~F1~$=1.0$ under the deployment configuration alone by
$N\approx700$, with no retuning required, reproducing the R1/R2 asymmetry from Fig.~\ref{fig:f2}.

\emph{Results, noisy (Fig.~\ref{fig:si-hyperparam}B).} At $N=5{,}000$, introducing measurement
noise collapses the recovery regardless of configuration: the oracle-tuned threshold falls
from Soft~F1~$=1.0$ (noiseless) to $0.60$ at $\eta=0.02$ and $0.28$ at $\eta=0.05$; the
deployment threshold ($\lambda=0.05$, at $\alpha=0$) falls to $0.45$ and $0.14$ respectively. Critically, the small
\emph{fixed} threshold ($\lambda=0.01$), which helped in the noiseless leg by retaining
more coefficients, performs \emph{worse} than the deployment default under noise
($0.21$ at $\eta=0.02$, converging to the same floor as the default only at $\eta=0.05$, $0.13$ against $0.14$):
lowering the threshold also retains more noise- and mixing-induced wrong coefficients. This is an
empirical illustration of the threshold tradeoff, not a model of final STLSQ recovery.

\emph{Scope of the control.} The figure establishes that the R1 ceiling can move under
hyperparameter changes and that a threshold helpful without noise can be harmful once
measurement noise is propagated. The SINDy indicator itself is validated at the
deployment configuration, using the exact first-step ridge algebra and the propagated
noise process (Appendix~\ref{app:sindy}).
\section{Finite-candidate information and expected posterior support}\label{app:information-support}
\label{si:information-support}

The invariant-measure moment matrix describes which coefficient differences are visible
to the data. To translate that geometry into an information diagnostic without
claiming to model SINDy or PySR search, consider an ideal observer choosing among a
fixed finite set of candidate equations \(\{H_0,\ldots,H_K\}\). Let \(H_h\) be the
truth, let \(\pi_j\) be the prior mass of \(H_j\), and assume a Gaussian experiment
with common known covariance. For competitor \(j\), define the standardized residual
signature
\[
r_j=\sqrt{\mathcal B}\,M^{1/2}\Delta\xi_{hj},
\qquad
\Delta\xi_{hj}=\xi_j-\xi_h,
\]
where \(\mathcal B=N_{\mathrm{eff}}/\sigma_\varepsilon^2\) is the data budget,
\(N_{\mathrm{eff}}=N/\tau_{\mathrm{int}}\) is the number of effectively independent samples
(Appendices~\ref{si:finite-time} and~\ref{si:stats-hygiene}),
and \(M\) is the moment matrix of the candidate terms under the regime. Under the truth, with \(z\) standard
Gaussian,
\[
\log\frac{p(Y\mid H_j)}{p(Y\mid H_h)}
=z^\top r_j-\frac12\|r_j\|_2^2.
\]

\subsection{Expected support for the true hypothesis}

The central information object is
\begin{equation}
S_h
=
\mathbb E_{Y\mid H_h}\!\left[\Pr(H_h\mid Y)\right]
=
\mathbb E_z\!\left[
\frac{1}{
1+\sum_{j\ne h}(\pi_j/\pi_h)
\exp\!\left(z^\top r_j-\tfrac12\|r_j\|_2^2\right)}
\right].
\label{eq:expected-posterior-support}
\end{equation}
This is the expected posterior mass assigned to the true equation by the ideal
finite-candidate Gaussian observer. It is not a calibrated posterior output of SINDy
or PySR and is not called their recovery probability.

At zero excitation all likelihoods coincide, so the posterior equals the
prior and \(S_h=\pi_h\). If \(\pi_h=0\), the truth is excluded and its
posterior support remains zero for every finite dataset; if \(\pi_h=1\), it
remains one. For \(\pi_h>0\), define the prior-predictive distribution
\[
p_{\mathrm{prior}}(y)
=
\sum_{j=0}^{K}\pi_j\,p(y\mid H_j),
\qquad
Q_h(y)=\Pr(H_h\mid y).
\]
Then
\[
S_h
=
\mathbb E_{Y\mid H_h}[Q_h(Y)]
=
\frac{
\mathbb E_{Y\sim p_{\mathrm{prior}}}[Q_h(Y)^2]
}{\pi_h}.
\]
This identity makes the averaging measure explicit. For the correctly
specified Gaussian experiment, monotonicity under additional information
follows from Blackwell refinement and conditional Jensen.

\subsection{Pairwise evidence and joint confounder geometry}

For one competitor,
\begin{equation}
\gamma_{hj}
=
\frac{N_{\mathrm{eff}}}{\sigma_\varepsilon^2}
\Delta\xi_{hj}^\top M\Delta\xi_{hj}
=\|r_j\|_2^2.
\label{eq:information-rho}
\end{equation}
This is the noise-normalized squared separation accumulated from the trajectory; in
the equal-covariance Gaussian model it is twice the pairwise Kullback--Leibler
divergence. The standardized comparison has the form
\[
L_j=-\frac{\gamma_{hj}}2+\sqrt{\gamma_{hj}}\,Z_j.
\]
For two competitors, the comparison correlation is
\begin{equation}
c_{ij}
=
\frac{\Delta\xi_{hi}^\top M\Delta\xi_{hj}}
{\sqrt{
(\Delta\xi_{hi}^\top M\Delta\xi_{hi})
(\Delta\xi_{hj}^\top M\Delta\xi_{hj})}}.
\label{eq:comparison-correlation}
\end{equation}
It measures similarity of their residual signatures under the observed dynamical
measure. Independent confounders provide more independent opportunities for misleading
evidence; strongly correlated confounders are more redundant. Consequently, finite-data
support is not determined by \(\lambda_{\min}(M)\) or by one pairwise separation alone:
the full comparison geometry and the prior also matter.

\smallskip
\noindent\textbf{Remark (precision along the weakest direction).}
The same geometry limits the estimation of coefficients. Consider the Gaussian linear
model $y=\Theta_N\xi+\varepsilon$ with $\varepsilon\sim\mathcal N(0,\sigma_\varepsilon^2I_N)$,
and assume that $\widehat M_N=\Theta_N^\top\Theta_N/N$ is positive definite. The Fisher
information is $N\widehat M_N/\sigma_\varepsilon^2$, so the Cram\'{e}r--Rao inequality gives,
for every unbiased estimator $\widehat\xi$ and the unit eigenvector $u$ of
$\lambda_{\min}(\widehat M_N)$,
\begin{equation}
\operatorname{Var}\!\left(u^\top\widehat\xi\right)
\geq
\frac{\sigma_\varepsilon^2}{N\,\lambda_{\min}(\widehat M_N)}.
\label{eq:crb-direction}
\end{equation}
Estimating this combination of coefficients with standard deviation at most $\epsilon$
therefore requires $N\lambda_{\min}(\widehat M_N)\geq\sigma_\varepsilon^2/\epsilon^2$. This
is the continuous counterpart of Eq.~\eqref{eq:information-rho}: a competitor that differs
from the truth by $\delta u$ has $\gamma_{hj}=N\delta^2\lambda_{\min}(\widehat M_N)/\sigma_\varepsilon^2$
when $N_{\mathrm{eff}}=N$. As $N$ grows, $\widehat M_N\to M$ (Appendix~\ref{app:sindy}), so
the bound is set by $\lambda_{\min}(M)$ of the regime. It holds for unbiased estimators
under response noise only. SINDy and PySR are biased and, in our noise experiments, also
face noise in the candidate terms, so the bound describes the information in the data,
not the performance of either algorithm.

A hard maximum-a-posteriori selector over the same fixed candidate set is a valid
ideal benchmark, but its conditional accuracy for one fixed truth can be discontinuous
at zero excitation and need not vary monotonically with every information path. It is
therefore not used as the paper's central information visualization, and neither the
hard-MAP probability nor \(S_h\) is identified with the recovery probability of the
actual discovery algorithms.

\subsection{Physical example: gravity and a linear restoring force}

The distinction between predictive agreement and recovery of the governing mechanism
can be illustrated by a central-force problem. Let
\(\mathbf r(t)\in\mathbb R^2\) denote position and let \(r=\|\mathbf r\|_2\).
Consider the gravitational vector field
\[
f_{\mathrm g}(\mathbf r)
=
-\frac{\Gamma}{r^3}\mathbf r,
\]
where \(\Gamma>0\), and the isotropic linear restoring force
\[
f_{\mathrm s}(\mathbf r)
=
-\frac{\Gamma}{R^3}\mathbf r.
\]
On a circular orbit of radius \(R\), \(r(t)=R\) at every observed time, and therefore
\[
f_{\mathrm g}(\mathbf r(t))
=
-\frac{\Gamma}{R^3}\mathbf r(t)
=
f_{\mathrm s}(\mathbf r(t)).
\]
Consequently, if \(\mu_{\mathrm{orb}}\) is the invariant measure supported on this
orbit,
\[
D_{\mu_{\mathrm{orb}}}(f_{\mathrm g},f_{\mathrm s})
=
\int
\left\|
f_{\mathrm g}(\mathbf r)-f_{\mathrm s}(\mathbf r)
\right\|_2^2
\,d\mu_{\mathrm{orb}}(\mathbf r)
=
0.
\]
The two force laws differ away from the orbit, but trajectory observations confined
to the circular support cannot distinguish them. With matching initial conditions,
they generate the same circular motion even though one represents inverse-square
gravity and the other a linear spring.

The equivalence becomes approximate for a near-circular orbit. Write
\(r(t)=R[1+\epsilon(t)]\), where \(|\epsilon(t)|\leq\delta<1\). Pointwise along the
observed trajectory,
\[
\frac{
\left\|f_{\mathrm g}(\mathbf r(t))-f_{\mathrm s}(\mathbf r(t))\right\|_2
}{
\left\|f_{\mathrm g}(\mathbf r(t))\right\|_2
}
=
\left|
\left(\frac{r(t)}{R}\right)^3-1
\right|
\leq
3\delta+3\delta^2+\delta^3.
\]
Hence
\[
D_{\mu_{\mathrm{orb}}}(f_{\mathrm g},f_{\mathrm s})
\leq
\left(3\delta+3\delta^2+\delta^3\right)^2
\int
\left\|f_{\mathrm g}(\mathbf r)\right\|_2^2
\,d\mu_{\mathrm{orb}}(\mathbf r).
\]
The observable separation is therefore quadratic in the radial excursion to leading
order. In the Gaussian information experiment, the corresponding pairwise
information scales as
\[
\gamma_{\mathrm{gs}}
\propto
\frac{N_{\mathrm{eff}}}{\sigma_\varepsilon^2}
D_{\mu_{\mathrm{orb}}}(f_{\mathrm g},f_{\mathrm s})
=
O\!\left(
\frac{N_{\mathrm{eff}}\delta^2}{\sigma_\varepsilon^2}
\right).
\]
Thus the mechanisms are exactly indistinguishable on the circular orbit and can be
only weakly distinguishable at finite resolution on a sufficiently narrow
near-circular orbit. This is not a global equivalence of the force laws: it is an
observational equivalence, exact or approximate, created by the restricted region
sampled by the dynamics.

\subsection{Interpretation of the main-text information figure}

Figure~\ref{fig:info-support} evaluates Eq.~\eqref{eq:expected-posterior-support} in
three controlled settings. Panel A separates effective budget from ordered
dictionary excitation using
\(M(\tau)=\operatorname{diag}(\tau,\tau,\tau^2,\tau^2,\tau^2,\tau^2)\), where \(\tau\in[0,1]\) is the excitation level shown in the legend of panels A and C.
Panel B holds the pairwise evidence \(\gamma_{hj}\) equal across competitors and varies the equicorrelation \(c\)
(Eq.~\eqref{eq:comparison-correlation}, shown in the legend) of fifteen competitors, exposing the role of the full confounder geometry. Panel C varies the
prior mass on the truth while holding one wrong-biased competitor profile fixed. The
zero-excitation curve is the derived identity \(S_h=\pi_h\), and increased excitation
raises support without allowing an infinitesimal favorable prior to substitute for
data. SINDy and PySR add thresholding, regularization, stochastic search, and
model-selection effects beyond this information benchmark.

\section{Complete experimental protocol}
\label{app:experimental-details}

\subsection{Numerical integration and regime selection}

Both systems were integrated with the adaptive Runge--Kutta solver RK45 of SciPy
(\texttt{solve\_ivp}, relative tolerance $10^{-7}$, absolute tolerance $10^{-9}$) at a
fixed output sampling interval $\Delta t = 0.01\,\text{t.u.}$
Each trajectory was preceded by a transient discarded before recording
($T_{\rm burn} = 100\,\text{t.u.}$ for L84; $50\,\text{t.u.}$ for L96)
from a random initial condition $x_0 \sim \mathcal{N}(0,\,0.01\,I)$.

Dynamical regimes were identified by sweeping the forcing parameter $F$ over
100 linearly spaced values and estimating the maximum Lyapunov exponent (MLE,
$\lambda_1$) at each point via the Benettin QR method (see
\textit{Lyapunov exponents} below).
Grid points were classified as fixed-point (FP, $\lambda_1 < -0.01$),
limit-cycle (LC, $|\lambda_1| \leq 0.01$), or chaotic ($\lambda_1 > 0.02$);
the narrow transition band was excluded from data collection.

Candidate LC trajectories required additional verification using the full
Lyapunov spectrum: the second exponent $\lambda_2$ had to be sufficiently
negative ($\lambda_2 < -0.02$) to confirm true limit-cycle dynamics and
reject quasi-periodic tori, which share $\lambda_1 \approx 0$ but have
$\lambda_2 \approx 0$ as well.
Trajectories passing this spectral check were then assigned to one of two
LC sub-regimes by attractor amplitude (the mean state standard deviation
$\sigma_x$) rather than by any Lyapunov exponent, since $\lambda_1$
carries no discriminating information within the LC band.
The natural bimodality of the $\sigma_x$ distribution (reflecting two
attractor branches) was resolved by splitting at the largest gap, yielding a
small-amplitude sub-regime (R1) and a large-amplitude one (R2).

The five chaotic sub-regimes (R3--R7) were defined by non-overlapping routing
windows in $\lambda_1$ and ordered by increasing MLE.
Table~\ref{tab:regime-reference-values} lists the pinned (reference) $F$ value realizing each fixed-point and chaotic regime, and the amplitude range defining each limit-cycle sub-regime, for both L84 and L96; the forcing of the individual trajectories can differ from it (see \textit{Trajectory data and library construction}).

\subsection{Trajectory data and library construction}

For each regime, five independent trajectory slots were collected, each
starting from a fresh random initial condition after burn-in.
Each slot comprises a training pool of $N_{\rm pool} = 10{,}000$ uniformly
spaced snapshots ($T_{\rm pool} = 100\,\text{t.u.}$) followed by a held-out
test block of $N_{\rm test} = 2{,}000$ snapshots from the same continuous
trajectory.

SINDy was evaluated against a degree-3 polynomial library with a
constant term, yielding $p = 20$ candidate features for L84 ($n=3$ dimensions)
and $p = 56$ for L96 ($n=5$ dimensions).
Library columns were not normalized before regression; the ground-truth coefficients
of both systems are $\mathcal{O}(1)$ and no column scaling was required.
PySR's analogous configuration, an operator set rather than a fixed library,
is described in its own subsection below.
Exact time derivatives at each snapshot were computed analytically from the
right-hand side evaluated at the recorded state.

\emph{Ground-truth forcing.} The forcing $F$ enters the L84 right-hand side through the constant term
$0.25\,F$ of $\dot x_0$ (and every L96 equation through $+F$), so the ground-truth expressions used for scoring, and
the exact derivatives of the noiseless experiments, were built from the forcing $F$ of the specific trajectory that was
fitted, read from the stored data for each (regime, slot), and not from the regime reference value $F_{\rm ref}$
of Table~\ref{tab:regime-reference-values}. For L84 the two differ frequently: 22 of the 40 trajectories (8 regimes $\times$ 5 slots) have
$F\neq F_{\rm ref}$. Regimes R1, R2, R4 and R5 contain slots with different $F$
(R1: $3.06$--$4.45$; R2: $4.91$--$5.84$; R4: $8.05$--$8.52$; R5: $7.93$--$8.05$), R6 and R7 use single values
($8.63$ and $7.70$) that differ from $F_{\rm ref}$ ($8.75$ and $7.59$), and only R0 and R3 use $F=F_{\rm ref}$
throughout. An L84 regime is therefore a class of trajectories sharing a dynamical type, not a single forcing value.
In L96 every slot of a regime uses $F=F_{\rm ref}$.

\subsection{Derivative estimation}

In the data-volume and prior-quality experiments, derivatives were evaluated directly from the governing equations, without noise.
In the noise experiments, which use regimes R1, R2, R3, and R7, additive Gaussian noise was independently injected into each state dimension as $\tilde{x}_d = x_d + \varepsilon_d$, where $\varepsilon_d \sim \mathcal{N}(0,\,\eta^2\,\sigma_{x_d}^2)$, $\sigma_{x_d}$ is the standard deviation of the clean training trajectory for dimension $d$, and $\eta \in \{0, 0.01, 0.02, 0.05, 0.10, 0.15, 0.20\}$ denotes the noise ratio.

The finite-difference (FD) track estimated derivatives via second-order central differencing applied to the noisy state sequence. 
The Savitzky--Golay (SG) track fitted a degree-3 polynomial over a sliding window of 11 time steps; both the smoothed state $\tilde{X}$ and the derivative $\dot{\tilde{X}}$ were extracted from the same polynomial fit, maintaining feature and target consistency.

\subsection{Moment matrix and \texorpdfstring{$\sigma_{\min}$}{sigma-min}}

The moment matrix $M$ and its empirical estimator
$\hat{M} = (1/N)\,\Theta^\top\Theta$ quantify the excitation of a specified
dictionary under the dynamical regime. For the full-library analyses, $\hat{M}$ was
estimated by evaluating the degree-3 polynomial dictionary on the same state
observations used for regression in each experimental cell, yielding the design
matrix $\Theta$.
For the moment-matrix figure (Fig.~\ref{fig:fMM}) and the data-volume
experiments, $\Theta$ was built from clean trajectories; $N$ equals $5{,}000$ for the
figure panels and equals $n_{\rm samp}$ for each data-volume cell.
In the noise experiments, the SINDy indicator propagated the specified Gaussian
state-noise process through the same noisy or Savitzky--Golay-smoothed state sequence,
derivative estimator, polynomial library, ridge fit, and threshold used in the
corresponding experimental track. Thirty-two antithetic Monte Carlo replicates were
used to estimate the marginal mean and variance of each first-step ridge coefficient;
these moments determine the expected threshold counts in Eq.~\eqref{eq:fsindy}.
The PySR geometry and irreplaceability inputs were computed as described in Appendix~\ref{app:pysr}.
The minimum singular value $\sigma_{\min}(\Theta)$, related to the moment matrix
eigenvalue by $\lambda_{\min}(\hat{M}) = \sigma_{\min}^2(\Theta)/N$, was extracted via
full singular value decomposition.
All reported values are means over the five independent trajectory slots.

\subsection{Lyapunov exponents}

Lyapunov exponents were estimated using the Benettin QR
method~\citep{benettin1980lyapunov}. The coupled state--tangent system was
integrated using fixed-step RK4 with the analytical Jacobian and step size
$\Delta t_{\mathrm{Lyap}}=0.01$. A full $d\times d$ tangent frame was
propagated ($d=3$ for L84 and $d=5$ for L96) and orthonormalized by QR
decomposition every 10 integration steps, corresponding to $0.1$ time units.
The largest exponent $\lambda_1$ was used for the regime sweep, while the
second exponent $\lambda_2$ was additionally inspected to distinguish limit
cycles from quasiperiodic motion. Integration ran for
$T_{\rm Lyap}=5{,}000$ time units after a $1{,}000$-time-unit burn-in for L84,
and for $T_{\rm Lyap}=200$ time units after a $50$-time-unit burn-in for L96.
Sweep estimates were averaged over 10 independent initial conditions per
forcing value.

\subsection{SINDy}

SINDy was run using the PySINDy implementation~\citep{kaptanoglu2022pysindy} with
sequentially-thresholded least squares (STLSQ) as the optimizer.
The sparsity threshold was fixed at $\lambda = 0.05$ and the ridge regularization
at $\alpha = 0.05$ across all experiments; both values were fixed prior to any
regime-stratified analysis and not tuned per regime. Appendix~\ref{app:hyperparam-control}
shows how the R1 plateau changes with the threshold and ridge parameter.
The maximum number of STLSQ iterations was set to 1{,}000 across all experiments.
The SINDy indicator is evaluated with the same $\alpha$ and $\lambda$ that produced
the corresponding fit; Eq.~\eqref{eq:sindy-exact-ridge} shows where both enter the first ridge step.

For derivatives, SINDy used the exact values in the data-volume and prior-quality
experiments and the FD or SG estimates in the noise experiments
(\textit{Derivative estimation}).
For the prior-quality experiment, the SINDy feature library varied by prior level
using the same three-level scheme as PySR (null, overcomplete, oracle), constructed
per system so that the ground-truth terms are respectively absent,
embedded among distractors, or exactly and exclusively present. For L84, the null SINDy library contains eight cubic decoys selected from the
ten degree-3 monomials so that its cardinality matches the eight-feature oracle
library. The selected terms are
\[
x_0^3,\;x_1^3,\;x_2^3,\;x_0x_1^2,\;x_0x_2^2,\;
x_1^2x_2,\;x_1x_2^2,\;x_0x_1x_2,
\]
with $x_0^2x_1$ and $x_0^2x_2$ omitted. None of the selected decoys is a
ground-truth term.
The overcomplete library is the full degree-3 polynomial library with bias (20
features, containing all ground-truth terms among 12 distractors); and the oracle library
is the 8 exact ground-truth terms
($1,\,x_0,\,x_1,\,x_2,\,x_1^2,\,x_2^2,\,x_0x_1,\,x_0x_2$) and nothing else, so
regression under the oracle prior reduces to ordinary least squares over the ground-truth
support. The L96 prior levels follow the same construction principle applied to
its own right-hand side (Appendix~\ref{app:l96}). The oracle condition is by construction close to
unconstrained regression on the ground-truth terms: an intentional upper-anchor that
isolates coefficient estimation from term selection, not a realistic operating point.

Results for each experimental cell are the mean Soft~F1 over the five independent
trajectory slots.

\subsection{PySR}

PySR version 1.5.9~\citep{cranmer2023pysr} was run with 50 evolutionary
iterations and 40 parallel island populations. Final expressions were selected
using \texttt{model\_selection="best"}: among the expressions of the Pareto front whose
loss is within a factor $1.5$ of the lowest loss, PySR returns the one with the highest
score, where the score is the decrease in log-loss per unit increase in complexity
along the front. This setting was identical across all experiments.
The operator set for the data-volume and noise experiments comprised the binary
operators $\{+,\,-,\,\times\}$ with no unary operators and a maximum expression
complexity of 20 nodes.
For the prior-quality experiment the operator set varied by prior level: the null
prior used $\{+,\,-\}$ over pre-computed decoy columns, the same decoys as the SINDy null
library (listed above for L84 and in Appendix~\ref{app:l96} for L96), so no expression can
contain a true term; the
overcomplete prior used $\{+,\,-,\,\times\}$ with complexity up to 25; and the
oracle prior used $\{+,\,-\}$ over the pre-computed ground-truth term columns with
complexity capped at 9.
Parsimony and adaptive parsimony scaling were both set to zero in all experiments,
so selection pressure derived purely from fitness.
For each experimental cell, five independent data slots were each run with five
independent random initialisations of the evolutionary search, giving 25 runs per
cell; the reported Soft~F1 is the mean over all runs.

\subsection{Statistical analysis}

Uncertainty bands in Fig.~\ref{fig:f2} are standard errors of the mean across independent fits. The three state
dimensions of one fit are not independent replicates, so each fit is first averaged over its dimensions; the
replicates are the 5 trajectory slots (SINDy) and the slot--initialisation runs (PySR, up to 25 per cell).

Associations between mechanistic indicator scores and observed Soft~F1 were quantified
by the Spearman rank correlation $\rho$, chosen for robustness to the monotone
but nonlinear relationship between theoretical predictors and empirical performance.
For both SINDy and PySR, $\rho$ was computed separately for each output
dimension, correlating $\mathcal I_{\mathrm{SINDy}}$ (respectively
$\mathcal{F}_{\mathrm{PySR}}$) with Soft~F1 across the cells of each experiment.
The factorial design comprised $680$ cells per equation in the data-volume
experiment, $280$ in the noise experiment, and $120$ in prior quality for
each system. The reported value is the mean $\rho$ across dimensions. L84's three
structurally distinct equations show systematic per-dimension heterogeneity that this
averaging absorbs (Appendix~\ref{si:per-dimension}); L96's five cyclically identical equations do not.
For both indicators, uncertainty was assessed by a cluster bootstrap that
resamples whole trajectories (regime$\times$slot) with replacement ($5000$ replicates) and
recomputes the per-equation-averaged $\rho$ (Appendix~\ref{si:stats-hygiene}). Collapsing the design to one value per regime leaves at most eight points,
too few to establish the within-regime-type ordering for L84; this regime-level analysis
and the effective-sample-size estimates appear in Appendix~\ref{app:l96}.

\end{document}